\documentclass{article}

\PassOptionsToPackage{sort&compress,numbers}{natbib}

\usepackage[final]{neurips_2022}

\usepackage[utf8]{inputenc} 
\usepackage[T1]{fontenc}    
\usepackage{hyperref}       
\usepackage{url}            
\usepackage{booktabs}       
\usepackage{amsfonts}       
\usepackage{nicefrac}       
\usepackage{microtype}      
\usepackage{xcolor}         
\usepackage[normalem]{ulem}
\usepackage{tablefootnote} 
\usepackage{todonotes}

\newcommand{\Nc}{\mathbf{N}_{c}}
\newcommand{\Nm}{\mathbf{N}_{m}}
\newcommand{\dNc}{\dot{\mathbf{N}}_{c}}

\newcommand{\obs}[1]{\tilde{#1}} 
\newcommand{\pred}[1]{\hat{#1}} 
\newcommand{\sample}[1]{\mathring{#1}}

\usepackage{amsthm}
\newtheorem{theorem}{Theorem}

\title{Bayesian Spline Learning for Equation Discovery of Nonlinear Dynamics with Quantified Uncertainty}

\author{%
Luning Sun\thanks{Equal contribution}\\
  University of Notre Dame\\
  \texttt{lsun7@nd.edu}\\
  \And
  Daniel Zhengyu Huang\footnotemark[1]\\
  California Institute of Technology\\
  \texttt{dzhuang@caltech.edu}\\ 
  \AND
  Hao Sun \\
  Renmin University of China  \\
  \texttt{haosun@ruc.edu.cn} \\
  \And
  Jian-Xun Wang \thanks{Corresponding author}\\
 	 University of Notre Dame \\
  	\texttt{jwang33@nd.edu} \\
}
\usepackage{amsmath}

\DeclareMathOperator*{\argmin}{arg\,min}
\usepackage{amsthm}
\usepackage{graphicx}
\usepackage[font=normalsize,labelfont=bf]{caption}
\usepackage[font=normalsize]{subfig}
\usepackage{float}
\usepackage[ruled,vlined]{algorithm2e}
\usepackage{algpseudocode}

\makeatletter
\newcommand*\bigcdot{\mathpalette\bigcdot@{.5}}
\newcommand*\bigcdot@[2]{\mathbin{\vcenter{\hbox{\scalebox{#2}{$\m@th#1\bullet$}}}}}
\makeatother

\begin{document}

\maketitle

\begin{abstract}
Nonlinear dynamics are ubiquitous in science and engineering applications, but the physics of most complex systems is far from being fully understood. Discovering interpretable governing equations from measurement data can help us understand and predict the behavior of complex dynamic systems. Although extensive work has recently been done in this field, robustly distilling explicit model forms from very sparse data with considerable noise remains intractable. Moreover, quantifying and propagating the uncertainty of the identified system from noisy data is challenging, and relevant literature is still limited. To bridge this gap, we develop a novel Bayesian spline learning framework to identify parsimonious governing equations of nonlinear (spatio)temporal dynamics from sparse, noisy data with quantified uncertainty. The proposed method utilizes spline basis to handle the data scarcity and measurement noise, upon which a group of derivatives can be accurately computed to form a library of candidate model terms. The equation residuals are used to inform the spline learning in a Bayesian manner, where approximate Bayesian uncertainty calibration techniques are employed to approximate posterior distributions of the trainable parameters. To promote the sparsity, an iterative sequential-threshold Bayesian learning approach is developed, using the alternative direction optimization strategy to systematically approximate L0 sparsity constraints. The proposed algorithm is evaluated on multiple nonlinear dynamical systems governed by canonical ordinary and partial differential equations, and the merit/superiority of the proposed method is demonstrated by comparison with state-of-the-art methods.
\end{abstract}

\section{Introduction}
\label{sec:intro}

In the realm of science and engineering, dynamical systems are ubiquitous. However, in actual circumstances, the governing equations behind complicated dynamics may not be completely understood, preventing researchers from developing first-principled models. On the other hand, the ever-increasing data availability opens up new avenues for scientists to identify predictive models from enormous observation data, a process known as \emph{system identification} (SI).
Recent advances in deep learning have prompted the rapid development of powerful SI models for high-dimensional problems using deep neural networks (DNNs). Many DNN-based SI models have been proposed to learn differential operators for complex (spatio)temporal physics from data and shown good potential in terms of data reproduction and state prediction~\cite{pfaff2020learning,lu2021learning,han2022predicting}.  
However, deep learning models usually lack interpretability and are difficult to comprehend. Furthermore, it is questionable in terms of generalizability when compared to first-principle models, as such black-box DNN models provide less insight into the underlying processes. 

Instead of identifying a black-box model, we focus on extracting analytical equation forms from data, which has higher interpretability and has the potential to advance our knowledge of unknown physics. 
The sparse identification of non-linear dynamic (SINDy) algorithm~\cite{brunton2016discovering} is an excellent development along this path. The central idea is to use sparse linear regression to uncover parsimonious governing equations from a dictionary of basis functions constructed by data, where the sparsity is promoted by pruning out redundant terms based on certain specified thresholds~\cite{rudy2017data}.
SINDy, although showing great promise, faces several challenges: (1) it heavily relies on high-quality data to extract derivative information, which is typically based on finite difference (FD) methods, making it impossible to handle incomplete, scarce, or noisy data; (2) it is formulated in a deterministic fashion and cannot account for the uncertainties from multiple sources, which is critical for real-world applications where data is frequently corrupted and can be very sparse. 
%
In the past few years, the SINDy framework has been further improved in various aspects to address these challenges, e.g., enhancing the library~\cite{chu2020discovering} or using deep learning for denoising and derivative computation by fitting the noisy data in a decoupled~\cite{rudy2019deep,wang2020denoising,wu2021semi} or coupled manner~\cite{long2019pde,kim2020integration,corbetta2020application,chen2021physics,sun2021physics}. For uncertainty quantification, the dictionary-based equation discovery algorithms have been recently extended to Bayesian settings~\cite{zhang2018robust, zhang2019robust,hirsh2021sparsifying}, based on the idea of sparse Bayesian learning pioneered by Tipping and co-workers~\cite{tipping2001sparse,tipping2003fast,bishop2013variational,faul2001variational,faul2002analysis}. 
Despite recent progress and extensive work in this field, reliably distilling explicit equation forms from very sparse data with significant noise remains an unsolved challenge. There are still significant gaps in handling data scarcity and noise, quantifying and reducing multi-source uncertainties, and promoting sparsity, which most existing equation discovery techniques struggle to simultaneously address. To this end, we propose a novel Bayesian Spline Learning (BSL) approach to identify parsimonious ordinal/partial differential equations (ODEs/PDEs) from sparse and noisy measurements; meanwhile, the associated uncertainties are quantified.
To deal with data scarcity and measurement noise, the proposed BSL uses a spline basis, on which a collection of derivatives can be reliably computed for the library construction. The posterior distributions of spline-based model parameters are approximated by a stochastic gradient descent (SGD) trajectory-based training scheme, where the first moment of SGD iterations is computed by stochastic weight averaging approach~\cite{maddox2019simple}.
The proposed BSL approach is effective in two aspects. On the one hand, the spline-based representation can help to interpolate locally the spatiotemporal field and perform differentiation analytically. As a result, it considerably enhances learning efficiency in cases with sparse and noisy data. The posterior distributions of spline, library, and equation coefficients are estimated simultaneously, without adding too much overhead to the training process.
Besides the measurement uncertainty, the model-form uncertainty is also obtained in the proposed method, which can be used for downstream Bayesian data assimilation, where online data is assimilated to improve the predictability of the identified system for chaotic scenarios.\footnote{The code will be available at \url{https://github.com/luningsun/SplineLearningEquation}}  

\section{Related Work}
\label{sec:related}
The data in real-world circumstances is frequently sparse in spatial/temporal domains and may contain considerable noise, posing significant challenges to SINDy or its variants~\cite{brunton2016discovering,rudy2017data,quade2018sparse,champion2019data,champion2019discovery,chu2020discovering,zhang2019convergence,zhang2019robust}. Deep learning (DL) has been leveraged as a superb interpolator for concurrently generating metadata and smoothing high-frequency noise~\cite{tancik2020fourier}, effectively improving the performance in identifying equations from imperfect measurements~\cite{long2019pde,kim2020integration,corbetta2020application,chen2021physics,sun2021physics,rudy2019deep}. 
For example, automatic differentiation (AD) is effective for computing derivatives analytically from sparse, noisy data using a point-wise multi-layer perceptron (MLP)~\cite{raissi2019physics,both2021deepmod,chen2021physics}. However, it is difficult to impose locality constraints in the point-wise formulation. Wandel et al.~\cite{wandel2021spline} demonstrated the merit of employing a spline basis for analytically calculating derivatives while enforcing the locality of spatiotemporal fields. Owing to the superiority of spline-based differentiation compared to numerical discretization, physics-informed spline networks remarkably outperform point-wise physics-informed neural networks (PINN) in the context of solving PDEs~\cite{wandel2021spline} and discovering ODEs~\cite{sun2021physics}. 

However, these models fail to simultaneously deal with ODE and PDE systems when the data is sparse and substantially corrupted.
For the PDE datasets, for example, the PINN-based sparse regression (PINN-SR)~\cite{chen2021physics} works admirably, but it fails to converge on ODE datasets. On the other hand, algorithms such as physics-informed spline learning (PiSL)~\cite{sun2021physics} that performs very well for ODE discovery cannot handle PDE problems due to the limitation of the B-spline basis adopted. More importantly, all these SOTA sparse learning algorithms are formulated in deterministic settings and uncertainties introduced from data/library imperfection cannot be quantified.  
For inverse problems like equation discovery, it is natural to use Bayesian framework to quantify and analyze the prediction uncertainty. To enable the posterior computation for high-dimensional trainable parameters that are intractable by traditional Bayesian inference, people have resorted to various approximation strategies, e.g., variational inference~\cite{blundell2015weight,liu2016stein,blei2017variational}, Monte Carlo dropout~\cite{gal2016dropout}, Bayes by backprop~\cite{Ebrahimi2020Uncertainty-guided}, Laplace approximation~\cite{ritter2018scalable}, and deep ensemble approaches~\cite{wilson2020bayesian}.    
Although these techniques have had great success, training may still be challenging, and costs will rise dramatically as the problems become more sophisticated. 
Alternatively, we chose to employ an SGD trajectory-based algorithm, Stochastic Weight Averaging-Gaussian (SWAG)~\cite{maddox2019simple}, where the information in the SGD trajectory is exploited to approximate the posterior. As demonstrated empirically, SWAC is well scalable to high-dimensional problems and can accurately estimate uncertainty across many different Bayesian learning tasks~\cite{maddox2019simple,abdar2021review}.

Data assimilation (DA) has been widely used in numerical weather prediction (NWP) by fusing online sensing data into a predictive model for nonlinear dynamics forecast. People have recently integrate deep learning into DA to improve online prediction performance~\cite{frerix2021variational,gronquist2021deep,mccabe2021learning}. However, the predictive model in DA is assumed to be known \emph{a priori}, and uncertainties from the model and observations, which are required for DA, are usually hard to obtain.  
In our framework, the predictive model can be unknown \emph{a priori} and will be identified to assimilate additional data for online forecasting. Moreover, instead of arbitrarily guessing the observation and model-form errors, this information can be learned in the proposed BSL framework, as the data and model-form uncertainties are quantified. The UQ capability of BSL naturally integrates the equation discovery with Bayesian DA techniques.

The main contributions of this work are three-fold: (1) we extended spline learning for sparse equation discovery of spatiotemporal physics governed by ODEs or PDEs; (2) we developed sparsity-promoting Bayesian learning for UQ tasks; and (3) the proposed BSL framework can seamlessly integrate with Bayesian data assimilation techniques to improve online dynamics forecasting.

\section{Methodology}
\label{sec:meth}
Let us consider a dynamical system, which is governed by a parametric ODE/PDE system in the following general form
\begin{equation}
\label{eq:dynamicalsystem}
\frac{d \mathbf{u}}{d t} = \mathcal{F}(\mathbf{u}),
\end{equation}
where $\mathbf{u}$ denotes the state vector, for ODE systems, $\mathbf{u} = [u_1(t), u_2(t), \cdots, u_d(t)]^T\in\mathbb{R}^{d}$ depends only on time $t$, for PDE systems, $\mathbf{u(x,}$ $t) = [u_1(\mathbf{x},t), u_2(\mathbf{x},t),\cdots, u_d(\mathbf{x},t)]^d\in\mathbb{R}^{d}$ depends on both time $t$ and space $\mathbf{x}$, and $\mathcal{F}: \mathbb{R}^d \to \mathbb{R}^d$ represents unknown nonlinear functions. The states are observed at discrete times $\{t_i\}_{i=1}^{n}$ and at spatial locations $\{\mathbf{x}_j\}_{j=1}^{s}$ for PDE systems. Let $\obs{\mathbf{u}}$ denote the noisy observation vector, the observation set is $\obs{\mathbf{U}} = \{\obs{\mathbf{u}}(t_1), \obs{\mathbf{u}}(t_2), \cdots, \obs{\mathbf{u}}(t_n)\}^T \in \mathbb{R}^{n \times d}$ and $\obs{\mathbf{U}} = \{\obs{\mathbf{u}}(\mathbf{x}_1,t_1), \obs{\mathbf{u}}(\mathbf{x}_2, t_1), \cdots, \obs{\mathbf{u}}(\mathbf{x}_s, t_n)\}^T \in \mathbb{R}^{(n \times s)\times d}$ for ODE/PDE systems, respectively.
Our goal here is to explicitly discover the parsimonious form of $\mathcal{F}(\cdot)$ from a library of candidate basis functions and quantify the associated uncertainty given noisy observation data.

 \begin{figure}
  \centering
  \includegraphics[width=0.95\textwidth]{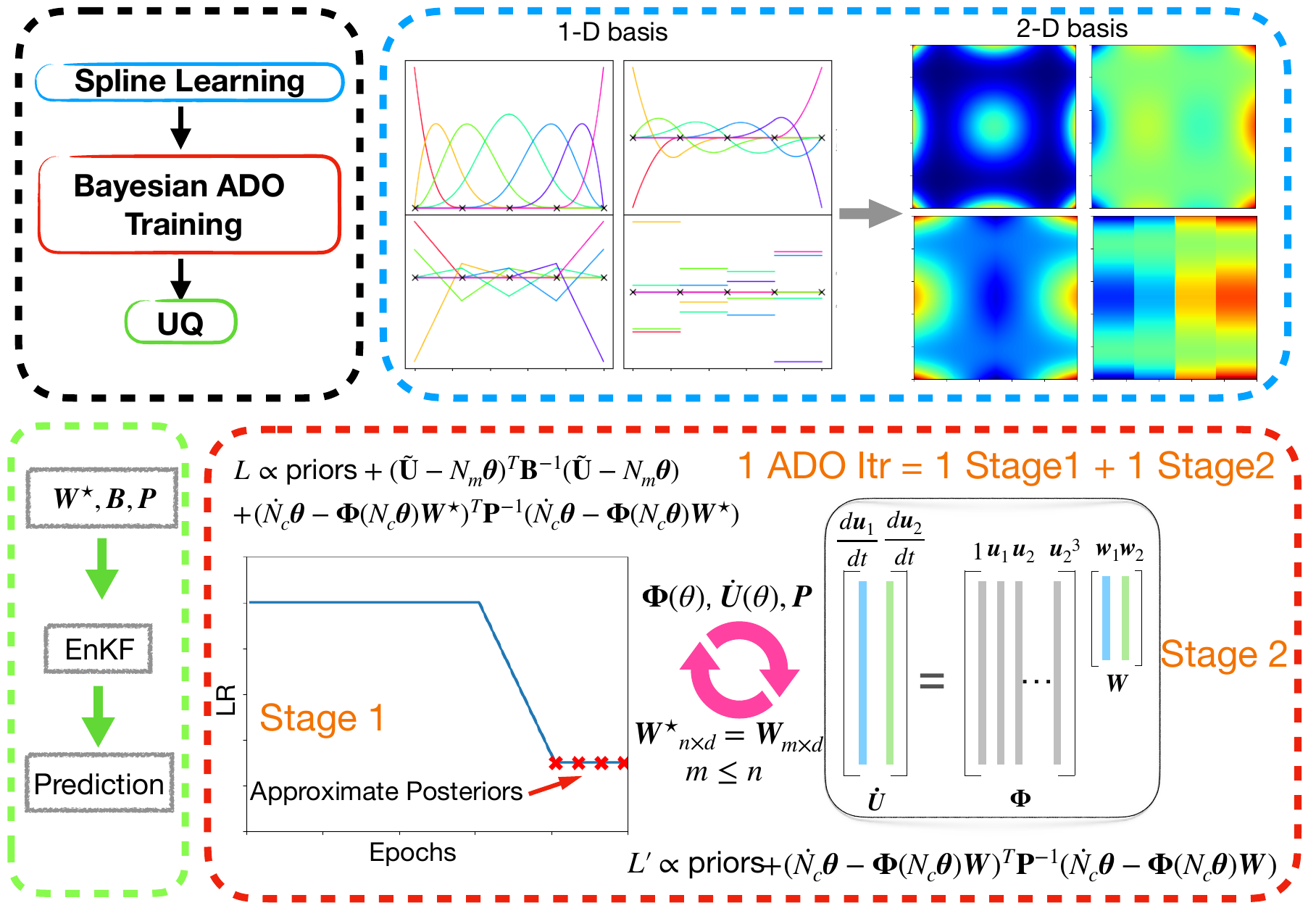}
  \caption{Overview of method. Black box: overall work flow. Blue box: a demo for spline basis with derivatives in 1D and 2D cases. Red box: a sketch for the Bayesian ADO training process. Green box: the data assimilation enhanced uncertainty quantification.}
  \label{fig:method}
  \vspace{0pt}
 \end{figure}
 
\subsection{Overview}
\label{ssec:SINDy}
Given the dataset $\obs{\mathbf{U}}$, the SI problem can be solved by sparse regression techniques with a predefined library $\mathbf{\Phi}(\mathbf{u})$ of $m$ basis functions,
\begin{equation}
\label{eq:lib}
\mathbf{\Phi}(\mathbf{u}) = [\phi_1(\mathbf{u}), \phi_2(\mathbf{u}), \cdots, \phi_m(\mathbf{u})] \in \mathbb{R}^m,
\end{equation}
where $\phi_i: \mathbb{R}^d \to \mathbb{R}, 1\leq i \leq m$ denotes a basis function, which, for instance, can be the polynomial or trigonometric function. Hence, the matrix of library terms evaluated at observed states is defined as,
\begin{equation}
  \begin{aligned}
\label{eq:matrixlib}
&\textrm{ODE :} &&\mathbf{\Phi}(\obs{\mathbf{U}}) = \bigg[\mathbf{\Phi}\big(\obs{\mathbf{u}}(t_1)\big)^T, \mathbf{\Phi}\big(\obs{\mathbf{u}}(t_2)\big)^T, \cdots, \mathbf{\Phi}\big(\obs{\mathbf{u}}(t_n)\big)^T\bigg]^T \in \mathbb{R}^{n \times m},\\
&\textrm{PDE :} && \mathbf{\Phi}(\obs{\mathbf{U}}) = \bigg[\mathbf{\Phi}\big(\obs{\mathbf{u}}(\mathbf{x}_1,t_1)\big)^T, \mathbf{\Phi}\big(\obs{\mathbf{u}}(\mathbf{x}_{2},t_1)\big)^T, \cdots, \mathbf{\Phi}\big(\obs{\mathbf{u}}(\mathbf{x}_s,t_n)\big)^T\bigg]^T \in \mathbb{R}^{(n \times s) \times m}.
 \end{aligned}
\end{equation}

Given the library, the overview of the BSL framework is shown in Fig.~\ref{fig:method}.
Firstly, a spline-based model is constructed to represent state variables and their derivatives by denoising the observation data, which is illustrated in the blue box and discussed in Sec.\ref{subsec:spline}. 
Then the spline-based model and sparse regression are trained simultaneously by alternating direction optimization (ADO), as shown in the red box. Specifically, a single ADO iteration contains two sub-processes, where sub-process I trains the spline-based model using log posterior loss and passes the trainable parameters and the noise estimations to sub-process II, which  uses a Bayesian SINDy-like method to prune out redundant terms in the library as defined in Eq.~\ref{eq:matrixlib}, and then updates the number of relevant terms in the training loss for sub-process I. After several ADO iterations, the parsimonious form of the governing equations and the posterior distribution for the coefficients will be estimated. More details about the ADO iterations in the Bayesian framework are shown in Sec.~\ref{subsec:BayesADO} and {Appendix~\ref{subsec:algs}}. 
Finally, with the estimated posterior, the predictive uncertainty can be quantified by evaluating the identified system with an ensemble of parameters. To further improve the prediction capability, especially for chaotic systems, we propose to leverage data assimilation techniques, which is shown in the green box and discussed in Sec.\ref{subsec:DA} and {Appendix~\ref{subsec:ODE1}}.

\subsection{Spline-based learning}
\label{subsec:spline}
Several previous works have already shown the potential of spline-based learning and demonstrated the advantages compared with the classical DL structures, e.g.,(MLP, CNN)~\cite{sun2021physics,wandel2021spline,fey2018splinecnn}. Therefore, we use this structure to smooth the solution fields based on the noisy measurement and then identify the underlying PDE/ODE and calculate the corresponding derivatives from the spline-reconstructed fields. The B-spline curves, based on the De Boor's algorithm, are defined in a recursive way as:
\begin{align}
\label{eq:spline1D}
\begin{aligned}
  N_{s,0}(t)&=\begin{cases}
    1, & \text{if $\tau_s\leq t \leq \tau_{s+1}$}.\\
    0, & \text{otherwise}.
  \end{cases}\\
  N_{s,k}(t)&=\frac{t-\tau_{s}}{\tau_{s+k}-\tau_{s}}N_{s,k-1}(t)+\frac{\tau_{s+k+1}-t}{\tau_{s+k+1}-\tau_{s+1}}N_{s+1,k-1}(t),
 \end{aligned}
\end{align}
where $\tau_{s}$ is the location of knots, $k$ is the degree of polynomial. When $k=3$, it is the well-used Cubic-B Spline curve. With the defined basis, the spline interpolation can be write as $y(t)=\Sigma_{s=0}^{r+k-1}N_{s,k}(t)\theta_s$. And the number of control points {(trainable weights)} $\boldsymbol{\theta} \in \mathbb{R}^{r+k}$ is chosen empirically. It can be proved that the derivative of $p$ order B-spline basis is a function of $p-1$ order B-spline, written as:
\begin{equation}
\label{eq:splinederiv1D}
\frac{d}{dt}N_{s,k}(t) = \frac{k}{\tau_{s+k}-\tau_{s}}N_{s,k-1}(t)-\frac{k}{\tau_{s+k+1}-\tau_{s+1}}N_{s+1,k-1}(t).
\end{equation} 
The proof of Eq.~\ref{eq:splinederiv1D} is attached in
{Appendix~\ref{subsec:spderiv}}. The higher-order derivative can be calculated by recursively using Eq.~\ref{eq:splinederiv1D}. The analytical derivatives of spline basis are very beneficial for PDE discovery tasks since it always involves constructing library terms containing high order derivatives. With a proper order $p$, the first $p$ derivatives are accurate, and there is no error introduced during the derivation, as opposed to using the numerical methods to approximate derivatives. The spline function and its derivatives are defined in a one-dimensional scenario. It is straightforward to extend to n-dimension by direct using tensor-product. For example, for a two-dimensional problem with spatial-temporal fields, the basis can be defined as
\begin{equation}
    N_{s_1,s_2}^{k_1,k_2}(t_i,x_j) = N_{s_1,k_1}(t_i)N_{s_2,k_2}(x_j). 
\end{equation}
Here the two-dimensional basis is denoted by a different style to write it compactly.
Similarly, the partial derivative for two-dimensional basis is defined as:
\begin{equation}
    \frac{\partial^{(q_1+q_2)}N_{s_1,s_2}^{k_1,k_2}}{\partial t^{(q_1)}\partial x^{(q_2)}}(t_i,x_j) = \frac{d^{(q_1)}N_{s_1,k_1}}{dt^{(q_1)}}(t_i)\frac{d^{(q_2)}N_{s_2,k_2}}{dx^{(q_2)}}(x_j).
\end{equation}
With the definition of sparse system identification and spline reconstruction, the whole spline learning can be stated as follows: given noisy measurement data $\obs{\mathbf{U}}$, find the best sets of weights $\boldsymbol{\theta}$ and $\mathbf{W}$ 
so that data fitting loss and the weakly physics-informed loss can be minimized under sparsity constraints
, as shown in Eq.~\ref{eq:detLoss}:
\begin{equation}
\label{eq:detLoss}
\begin{aligned}
\{\boldsymbol{\theta},\mathbf{W}\} 
&= \argmin_{\boldsymbol{\theta}',\mathbf{W}'} \frac{1}{\mathcal{N}_m}||\Nm\boldsymbol{\theta}' - \obs{\mathbf{U}}||_{L_2} + \frac{1}{\mathcal{N}_c}||\mathbf{\Phi}(\Nc\boldsymbol{\theta}' ) \mathbf{W}' - \dNc\boldsymbol{\theta}'||_{L_2} + \lambda||\mathbf{W}'||_{L_\alpha}.\\
\end{aligned}
\end{equation}
Here, $\Nm$ and $\Nc$ denote the spline basis matrices evaluated at measurement and collocation locations. Moreover, $\mathcal{N}_m$ and $\mathcal{N}_c$ are numbers of measurement data and collocation points. Furthermore, the alternating direction optimization (ADO) shown in previous works~\cite{chen2021physics,sun2021physics} can be adopted to minimize the loss function efficiently, and the details are attached in {Appendix~\ref{subsec:algs}}.

\subsection{Sparse system identification in Bayesian formulation}
\label{subsec:BayesADO}
Bayesian methods provide a natural probabilistic representation of uncertainty, which is crucial for model predictions. System identification from noisy and sparse measurements generally contains two types of errors (similar to the hidden Markov model):
(1) Observation error, where the data is noisy, and the smoothed data is reconstructed through spline-based learning, as shown in Eq.~\ref{eq:reconstruct}
\begin{equation}
\label{eq:reconstruct}
\obs{\mathbf{U}} = \Nm\boldsymbol{\theta}'+\boldsymbol{\epsilon}_1,
\end{equation}
where $\boldsymbol{\epsilon}_1$ represents the observation error.
(2) Evolution error or model form error, 
since the discovered system cannot be exact due to library imperfection and needs to be reformulated as,
\begin{equation}
    \label{eq:sindyB}
    \dot{\mathbf{u}} = \boldsymbol{\Phi}(\mathbf{u})\boldsymbol{W'} + \boldsymbol{\epsilon}_2,
\end{equation}
where $\boldsymbol{\epsilon}_2$ represents model form error. And Eq.~\eqref{eq:sindyB} is evaluated on collocation points.

In this work, these error terms are modeled as zero-mean multivariate Gaussian random variables: $\boldsymbol{\epsilon_1}\sim \mathcal{N}(\mathbf{0},\mathbf{B})$ and $\boldsymbol{\epsilon_2} \sim \mathcal{N}(\mathbf{0}, \mathbf{P})$. And we further assume that these error covariance matrices $\mathbf{B}, \mathbf{P}$ are diagonal matrix with diagonal terms $\{b_k\}$ and $\{p_k\}$ with $1\leq k \leq d$ and they are learn-able parameters during training.

According to Bayes' rule, the posterior can be written as:
\begin{equation}
\label{eq:posterior}
    p(\boldsymbol{\theta},\mathbf{W}, \mathbf{B},\mathbf{P}|\obs{\mathbf{U}},\dot{\mathbf{U}})\propto  p(\boldsymbol{\theta},\mathbf{W}, \mathbf{B},\mathbf{P})p(\obs{\mathbf{U}},\dot{\mathbf{U}}|\mathbf{W},\boldsymbol{\theta}, \mathbf{B},\mathbf{P}).
\end{equation}\\
{Here $\dot{\mathbf{U}} = \dNc\boldsymbol{\theta}$ denotes the derivative estimation on all the collocation points.} 
The prior is further decomposed by
\begin{equation}
\label{eq:prior}
     p(\mathbf{W},\boldsymbol{\theta}, \mathbf{B},\mathbf{P})\propto p(\mathbf{W}|\alpha)p(\alpha)p(\boldsymbol{\theta}|\beta)p(\beta)p(\mathbf{B})p(\mathbf{P}).
\end{equation}
Currently, we specify the prior for the linear coefficient matrix as a zero mean Gaussian distribution $p(\mathbf{W}|\alpha) = \mathcal{N}(\mathbf{W}|0,\alpha^{-1}\mathbf{I})$ with the hyper prior as a Gamma distribution $p(\alpha) = \text{Gamma}(\alpha|a_0, b_0)$. Similarly, we also define the prior for the spline trainable parameters as zero mean Gaussian distribution with the hyper prior as another Gamma distribution $p(\mathbf{\boldsymbol{\theta}}|\beta) = \mathcal{N}(\boldsymbol{\theta}|0,\beta^{-1}\mathbf{I})$ and $p(\beta) = \text{Gamma}(\beta|a_1, b_1)$. To account for the data uncertainty and process uncertainty, the diagonal covariance matrices $\mathbf{B}$ and $\mathbf{P}$ are also set as learn-able during the training. And hence improper uniform priors are used for $\mathbf{B}$ and $\mathbf{P}$.\\
The likelihood consists of two parts, as:
\begin{equation}
\label{eq:likelihoodfunction2}
p(\obs{\mathbf{U}},\dot{\mathbf{U}}|\mathbf{W},\boldsymbol{\theta}, \mathbf{B},\mathbf{P}) = p(\obs{\mathbf{U}}|\boldsymbol{\theta}, \mathbf{B})p(\dot{\mathbf{U}}|\mathbf{W},\mathbf{P},\boldsymbol{\theta}),
\end{equation}
where
\begin{align}
p(\obs{\mathbf{U}}|\boldsymbol{\theta}, \mathbf{B})&\propto 
\exp\{-\frac{1}{2}(\obs{\mathbf{U}} - \Nm\boldsymbol{\theta})^T\mathbf{B}^{-1} (\obs{\mathbf{U}} - \Nm\boldsymbol{\theta})\},\\
\label{eq:llP}
p(\dot{\mathbf{U}}|\mathbf{W},\mathbf{P},\boldsymbol{\theta}) &\propto 
\exp\{-\frac{1}{2}( \dNc\boldsymbol{\theta} - \boldsymbol{\Phi}(\Nc\boldsymbol{\theta})\boldsymbol{W})^T\mathbf{P}^{-1} (\dNc\boldsymbol{\theta}- \boldsymbol{\Phi}(\Nc\boldsymbol{\theta})\boldsymbol{W})\}.
\end{align}

Traditional Bayesian sampling approaches~(i.e., Markov chain Monte Carlo methods) can be usually intractable and expensive, especially when the parameter dimensionality is high. Therefore, researchers tend to use alternative approximation approaches. Current work uses a stochastic gradient descent~(SGD) trajectory-based approach, Stochastic Weight Averaging Gaussian (SWAG)~\cite{maddox2019simple} algorithm to approximately sample from the posterior distribution. This method approximates the posterior by collecting the parameters near the loss plateau after a sufficient number of the training steps. To further reduce the inference cost, we construct a subspace by finding the PCA components of the SWAG trajectories~\cite{izmailov2020subspace} and then draw samples in the subspace instead. In terms of the loss function in the probabilistic model, we chose to maximize the log form of the posterior density Eq.~\ref{eq:posterior}, also by leveraging the ADO algorithm. The sub-process I minimizes the log density function defined by Eq.~\ref{eq:posterior}-\ref{eq:likelihoodfunction2} and the sub-process II adopts the Bayesian variants of SINDy  algorithm modified from ~\cite{tipping2003fast,hirsh2021sparsifying}. The likelihood function in sub-process II has the same form as in Eq.~\ref{eq:llP} but with a different sparsity promoting prior $p(\mathbf{W}|\mathbf{A}) = {\displaystyle \prod_{j=1}^{m} \mathcal{N}(\mathbf{W}_j|0, {\alpha'}_{j}^{-1})}$, where $\boldsymbol{A} = [\alpha'_{1}, \alpha'_{2}, ..., \alpha'_{m}]^{T}$. In a single ADO iteration, the sub-process I provides the updated $\theta$ and $\mathbf{P}$ to sub-process II for constructing the library terms. While the sub-process II shrinks the library terms and passes updated library/weight structure $\mathbf{W}$ back to sub-process I. The whole training requires multiple ADO iterations before it reaches the final balance, where no more terms will be pruned out in sub-process II. The detailed ADO algorithm is listed in {Appendix~\ref{subsec:algs}}. After obtaining the approximated posterior distribution, predicting uncertainty can be estimated by marginalizing out the model parameters.
 
\subsection{Data assimilation for enhanced predictability}
\label{subsec:DA}
For a chaotic system, one notorious problem is that a slight perturb in any model parameters can significantly influence the prediction. For example, in numerical weather prediction (NWP) tasks where the researcher always needs to predict the behavior of chaotic weather systems, various data assimilation (DA) techniques have been developed to assimilate the available data and the known equations. Kalman filter and its variants are successful mathematical tools for data assimilation. Our Bayesian formulation, providing the observation error covariance matrix $\mathbf{B}$ and evolution error covariance matrix $\mathbf{P}$, can be naturally incorporated with the Kalman filter frameworks to improve the predictability of chaotic systems. Current works choose the ensemble Kalman filter for the task, and more background can be found in {Appendix~\ref{subsec:algs}}.

\section{Experiment and Result}
\label{sec:result}
In this section, we first show the equation discovery and uncertainty quantification results for nonlinear ODE systems. We also show that incorporating DA techniques can improve the predicting ability for chaotic systems. Finally, we present the PDE discovery results for several canonical PDE systems.

\begin{figure}
   \centering
   \includegraphics[width=1.0\textwidth]{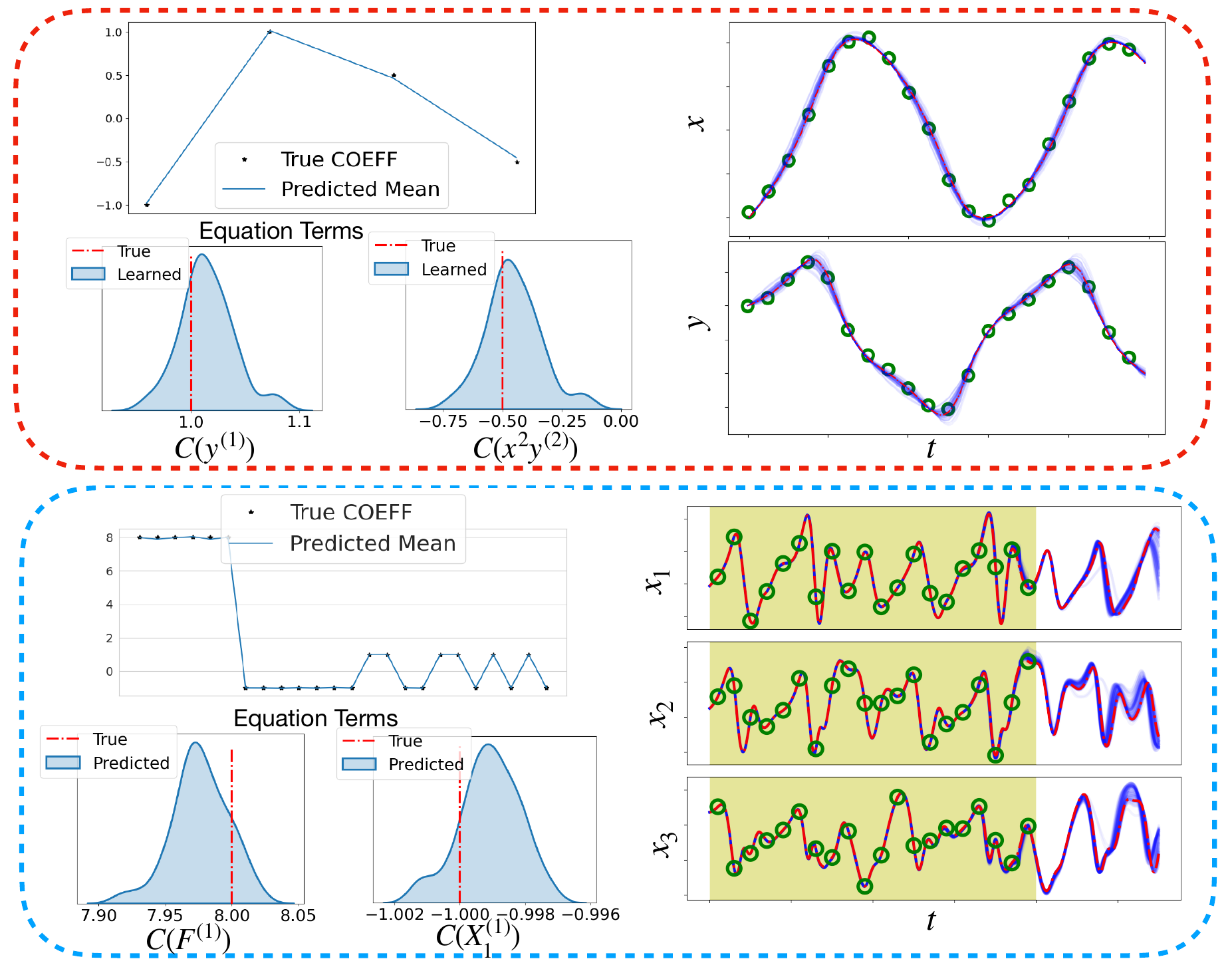}
   \caption{The discovery results for ODE systems; the red box shows results for the Van der Pol (VdP) system, and the blue box shows the result for Lorenz 96 system. The layout inside each box follows the rules below. Upper left sub-fig:  discovered mean for the relevant library terms; Lower left sub-figs: selected posterior distribution for the identified distributions; Right sub-figs: ensemble prediction plots for UQ. {For VdP system, the governing equation is $\frac{dx}{dt} = y$ and $\frac{dy}{dt} = -x-0.5x^2y+0.5y$. For the Lorenz96 system, the compact form of the governing equation is $\frac{dX_i}{dt} = (X_{i+1}-X_{i-2})X_{i-1}-X_i+F$ with periodic boundary conditions.}}
    
   \label{fig:vanDerPol}
 \end{figure}

The first pedagogical ODE example is the Van der Pol oscillator, which is defined as $\frac{dx}{dt} = y$ and $\frac{dy}{dt} = \mu(1-x^2)y-x$
with $\mu = 0.5$. The data is corrupted with $5\%$ noise, and the library consists of polynomials of state variables up to $3rd$ order. The parsimonious model structure can be correctly identified, and the result is shown in the upper red box of Fig.~\ref{fig:vanDerPol}. There are five sub-figures, which depict the library discovery, coefficient distribution, and the forward propagated UQ results. Specifically, the single upper left figure shows the mean of discovered coefficients (blue lines ${\color{blue}{\boldsymbol{-}}}$) and the truth equation coefficients (black stars ${\color{black}{\boldsymbol{*}}}$), where the horizontal axis represents term indices. For example, the Van der Pol system has four different terms, and the x-axis ranges from 1 to 4. The two sub-figures in the lower left part show probability density distributions (PDF) of two identified coefficients, where truth equation coefficients (red lines ${\color{red}{\boldsymbol{-}}}$) fall within
the confidence interval with high probability. The right sub-figures show the propagated ensemble results (blue lines ${\color{blue}{\boldsymbol{-}}}$) based on the discovered equations, measurement data (green circles $\color{green}{\mathbf{\circ}}$), and the true state trajectories of $x$ and $y$ (red dashed lines ${\color{red}{\boldsymbol{--}}}$). 
The result clearly shows that the ensemble predictions can cover the data, and the uncertainty range fluctuates around the truth state values.

The second ODE example is a chaotic system, Lorenz 96. It is a simplified mathematical model for atmospheric convection, defined as:
     $\frac{dX_i}{dt} = (X_{i+1}-X_{i-2})X_{i-1}-X_i+F$, $i=1,2,...n$ 
 with periodic boundary conditions $X_{-1}=X_{n-1}, X_{0} = X_{n}$ and $X_{n+1} = X_{1}$. We chose $n={4}$ in current case and $F = 8$ for the forcing terms. The measurement states variables are corrupted with $10\%$ Gaussian noise, and the library consists of polynomials of state variables up to $3rd$ order. The discovery result for these systems is shown in the lower blue box in Fig.~\ref{fig:vanDerPol} following the same layout as the Van der Pol systems. The upper left figure indicates that the sparsity structure can be identified, and the horizontal axis marks 24 parsimonious terms (4 terms for each state, 6 states in total) out of 84 library terms. The detailed PDF plots show that posterior distributions still cover truth values with high probability. Although the discovery result is quite accurate, the forward simulations of identified Lorenz 96 system would induce a large phase difference compared with truth trajectories due to the chaotic nature of the underlying system ({in Appendix~\ref{subsec:ODE1}}). To alleviate the chaotic behavior and improve the predictability, we seamlessly coupled the ensemble Kalman filter (EnKF), a classical DA technique, to assimilate noisy measurements with the identified system. The observation error matrix $\boldsymbol{B}$ and evolution error matrix $\boldsymbol{P}$ required for the EnKF scheme can be directly passed from the Bayesian spline learning framework. The right subplots show the DA results for the 3 out of 6 state variables to save space. The horizontal axis marks the evolving time, and the vertical axis represents the state variables. In each subplot, the green shaded region indicates the time interval with available noisy measurement data (marked by green circles ($\color{green}{\mathbf{\circ}}$)). And the red solid-line ($\color{red}{\boldsymbol{-}}$) is the true L96 states. Finally, the blue curves (${\color{blue}{\boldsymbol{-}}}$) are the ensemble predictions. It can be seen that the ensemble predictions inside the region with measurement data almost overlap with the true trajectory.
 Furthermore, the ensemble uncertainty grows more significantly in the extrapolation region, but the ensemble ranges still fluctuate around the true trajectory. Note that the prediction for the chaotic system with the DA process is much better than directly forward simulating the identified system, where the predictable interval is only about 1 second. More discovery results are attached in {Appendix ~\ref{subsec:ODE1}}.
 
  \begin{figure}{h}
  \centering
  \includegraphics[width=1.0\textwidth]{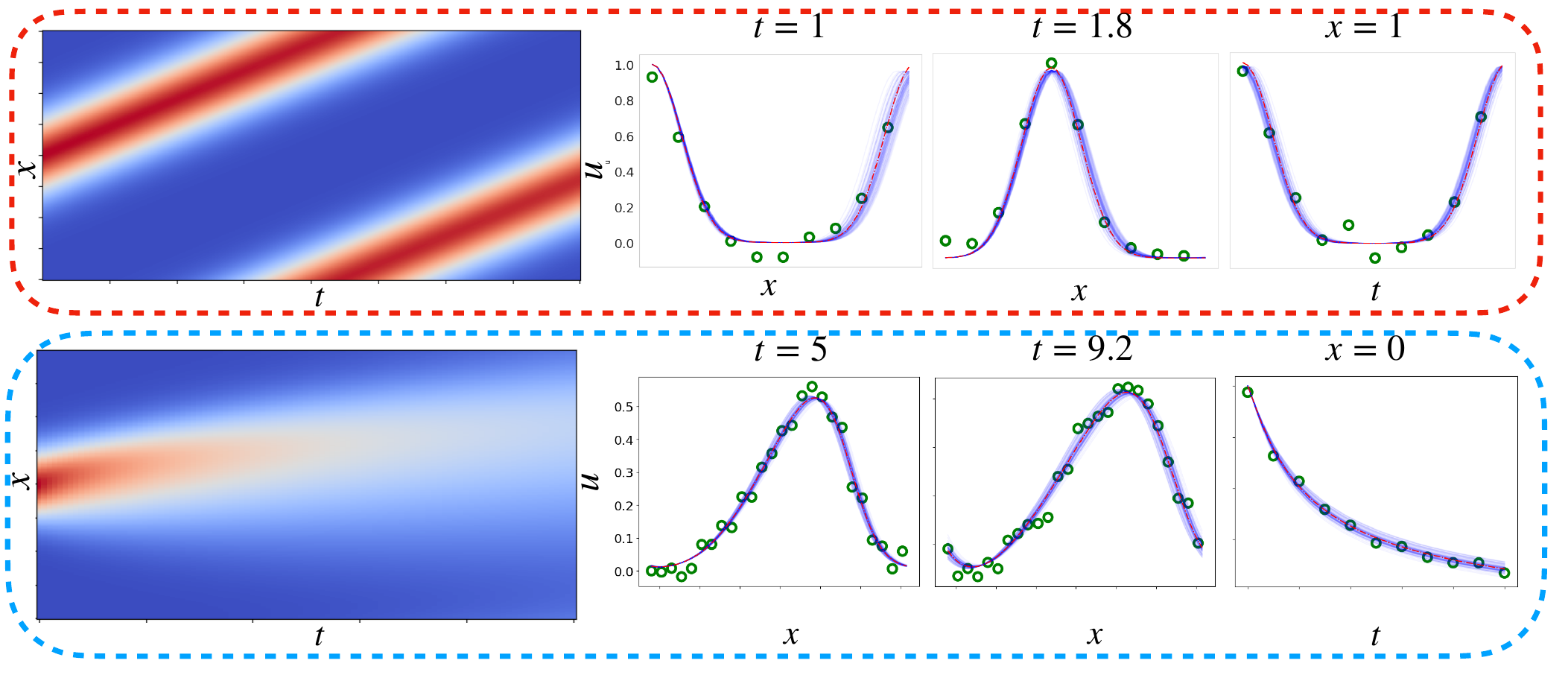}
  
  \caption{The discovery results for PDE systems; the red box shows results for the advection system, and the blue box shows the result for the Burgers' system. The layout inside each box follows the rules below. Leftmost sub-fig: true contour plot; Middle two sub-figs: the spatial results at different time $t$; Rightmost sub-figs: the temporal result at a fixed point $x$.}
  \label{fig:PDEf1}
 \end{figure}
Then, we use the proposed BSL framework to identify classical PDE systems and evaluate the performance of two of them in the main text. They include advection equation $\frac{\partial u}{\partial t} + \frac{\partial u}{\partial x} = 0$ and Burgers' equation $\frac{\partial u}{\partial t} + u\frac{\partial u}{\partial x} -0.5\frac{\partial^2u}{\partial x^2}= 0$. {The qualitative results are shown in Fig.~\ref{fig:PDEf1}} 
inside the red and blue boxes, respectively. Inside each box, the leftmost contour plot shows the true state value in the spatiotemporal field, and the right three sub-figures show the UQ results for different cross-sections from the left contour plot. 
The red lines (${\color{red}{\boldsymbol{-}}}$) represent the truth value, blue lines (${\color{blue}{\boldsymbol{-}}}$) are obtained by ensemble predictions, and green circles (${\color{green}{\mathbf{\circ}}}$) represent sparse and noisy measurements.
The prediction from the identified PDE system is accurate in the spatiotemporal field, and the ensemble fluctuates around the truth state variables.
  
Finally we benchmark our proposed method with several state-of-the-art discovery algorithms (PINN-SR~\cite{chen2021physics}, SINDy~\cite{rudy2017data} and {RVM~\cite{tipping1999relevance}}). The error metric is defined as:
 \begin{align}
 {\textbf{rmse}} &= \frac{||\mathbf{C}_{\rm Discovery}-\mathbf{C}_{\rm True}||_2}{||\mathbf{C}_{\rm True}||_2}\\
 {\mathbf{M_P}} &= {\frac{||\mathbf{C}_{\rm Discovery}\odot\mathbf{C}_{\rm True}||_0}{||\mathbf{C}_{\rm Discovery}||_0}}\\
{\mathbf{M_R}} &= {\frac{||\mathbf{C}_{\rm Discovery}\odot\mathbf{C}_{\rm True}||_0}{||\mathbf{C}_{\rm True}||_0}}
 \label{eq:PDEres2}
 \end{align}
 where $\mathbf{C}_{\rm Discovery}$ are the non-zero mean prediction from the posterior distribution and $\mathbf{C}_{\rm True}$ are the true coefficients of the governing equations. If the method fails to converge or cannot identify the correct parsimonious form, we will report the final result as {\textbf{Fail}}. The errors are scaled by $\times 10^{-3}$ to have a clear comparison. {Two additional metrics, precision $M_P$ and recall $M_R$, are also defined, where the $\odot$ represents element-wise product of vectors and the $l_0$ norm is the non-zero terms in a vector}. 
 It can be seen from Table~\ref{tab:ODEcp1} that our BSL method always performs best, when the noise is significant ($>5\%$), demonstrating its robustness to noise. 

 The PINN-SR can discover the PDE equation with corrupted data set, but it fails to predict accurate time trajectories for ODE systems. This is also reported in a relevant paper~\cite{wang2021long}, that the plain MLP structure is not satisfactory for predicting time series. The SINDy method can behave much better when large high-quality data exist. However, the SINDy method can easily fail when noise is significant. Our BSL requires fewer parameters and is easier to converge due to the enforcement of locality constraints by the spline basis model. Furthermore, the SINDy and RVM methods require much more data ($100\%$) to train but are still vulnerable to data noise. These benchmark cases show the potential of our BSL for equation discovery tasks. {The full table with more benchmark test cases can be found in Tab.~\ref{tab:FullODEcp1} in Appendix. We also apply the proposed method on a real-world dataset, as shown in Tab.~\ref{tab:discrealODE} and Tab.~\ref{tab:anarealODE} and discuss the effect of smoothing method in Tab.~\ref{tab:smoothmetric} and Tab.~\ref{tab:smooth_eq}.}
\begin{table}[h]
  \caption{ODE and PDE discovery comparison}
  \label{tab:ODEcp1}
  \vspace{6pt}
  \centering
  \footnotesize
  \begin{tabular}{lllllll}
    \toprule
    
    Name     &{$\textbf{rmse}$}($0\%$)    &{$\textbf{rmse}$}($1\%$) &{$\textbf{rmse}$} (large\tablefootnote{Large noise for different cases: Van der Pol: $5\%$, Lorenz 96: $10\%$, Advection: $20\%$, Burgers: $10\%$, Burgers' with source: $20\%$, Heat: $15\%$, Poisson: $5\%$})& {$\mathbf{M_P}$} &  {$\mathbf{M_R}$\tablefootnote{$\mathbf{M_P},\mathbf{M_R}$ are only reported for the largest noise cases}}&{Training Cost\tablefootnote{All cases are running on a Nvidia 2070 Ti GPU card}}\\
    \midrule
    \multicolumn{7}{c}{Van der Pol Oscillator}
    \\
    \cmidrule(r){1-7}
    \textbf{BSL(Ours)} & $\mathbf{0.2}$  &$2.82$&$\mathbf{18.04}$& {$\mathbf{1}$} &{$\mathbf{1}$}& {$\sim 133(+3)s$}   \\
    PINN-SR&$\text{Fail}$\tablefootnote{Fail means failure in discovery of the parsimonious ODE/PDE forms.}&$\text{Fail}$&$\text{Fail}$&{$0.214$}&{$0.75$}&{$\sim 1213s$}\\
    SINDy     & $1.0$       &$\mathbf{1.93}$&$\text{Fail}$& {$0.267$}&{$1.0$}& {$\sim 10s$} \\
    RVM     & {$1.0$}       &{$2.54$}&{$27.46$}& {$\mathbf{1}$}&{$\mathbf{1}$}& {$\sim 10s$} \\
    
    \toprule
    \multicolumn{7}{c}{Lorenz 96}                   \\
    \midrule
    \textbf{BSL(Ours)} & $\mathbf{0.269}$  &$1.47$&$\mathbf{13.0}$& {$\mathbf{1}$} &{$\mathbf{1}$}    &{$\sim1654(+438)s$}\\
    PINN-SR&$\text{Fail}$&$\text{Fail}$&$\text{Fail}$&{$0.5$}&{$0.22$}&{$\sim10788s$}\\
    SINDy     & $0.4$       &$0.64$&$\text{Fail}$ & {$0.75$} & {$\mathbf{1}$} & {$\sim 10s$}\\
    RVM     & {$0.4$}       &{$\mathbf{0.6}$}&{$49.7$}& {$\mathbf{1}$}&{$\mathbf{1}$}& {$\sim 25s$} \\
    \toprule
    \multicolumn{7}{c}{Advection Equation}                   \\
    \midrule
    \textbf{BSL(Ours)} & $\mathbf{0.26}$  &$\mathbf{1}$&$\mathbf{1.9}$& {$\mathbf{1}$} &{$\mathbf{1}$}    &{$\sim 946(+233)s$}\\
    PINN-SR & $5.9$ &$4.5$&$30.4$ &{$\mathbf{1}$} &{$\mathbf{1}$}&{$\sim 650s$}\\
    SINDy     & $2.3$       &$8.2$&$38.9$& ${\mathbf{1}}$&${\mathbf{1}}$& {$\sim 10s$} \\
    RVM     & {$0.77$}       &{$6.76$}&{$\text{Fail}$}& ${0.2}$&${\mathbf{1}}$& {$\sim 4s$} \\
    \toprule
    \multicolumn{7}{c}{Burgers' Equation}                   \\
    \midrule
    \textbf{BSL(Ours)} & $3.62$  &$4.13$&$\mathbf{6.38}$& ${\mathbf{1}}$ &${\mathbf{1}}$ & {$\sim 117(+74)s$}  \\
    PINN-SR & $10.2$ &$\mathbf{3.3}$&$10.3$ &${\mathbf{1}}$      &${\mathbf{1}}$&{$\sim 512s$}\\
    SINDy     & $0.826$       &Fail&Fail & ${\mathbf{1}}$ & ${0.5}$& {$\sim 10s$} \\
    RVM     & {$\mathbf{0.754}$}       &{$
    \text{Fail}$}&{$\text{Fail}$}& ${0.1429}$&${0.5}$& {$\sim 4s$} \\
    \bottomrule
  \end{tabular}
 \end{table}

\section{Discussion and Limitation}
\label{sec:conclusion}
In this work, we developed a novel Bayesian spline learning (BSL) framework for equation discovery from sparse and noisy datasets with quantified uncertainty. {The proposed framework significantly improves SI performance and contributes to the existing literature in the following aspects: Firstly, the use of spline basis enables us to accurately interpolate solution surfaces and analytically compute derivatives to form the candidate library, outperforming other benchmark methods based on finite difference (FD) or auto differentiation (AD), which either suffer from noisy/sparse data or overfitting issues.}  {Secondly, the proposed Bayesian learning formulation notably enhances the robustness for large data noise and sparsity, and meanwhile, quantifies the predictive uncertainty with minimum computational overhead.} {Moreover, a Bayesian sparsity-promoting ADO iteration strategy is proposed to promote sparsity and recover the parsimonious governing equation as well as the posterior distribution of its coefficients.}   
{Last but not least,} the Bayesian DA is also integrated into the BSL framework to improve the online predictability of the chaotic systems, {which can potentially benefit real-world tasks such as numerical weather forecasting}. The proposed framework is evaluated on discovering multiple canonical ODE and PDE systems, and great superiority has been demonstrated in comparison with state-of-the-art methods. 

Admittedly, this work still relies on a pre-defined library of candidate terms, and thus the identified system is largely limited to the functional space determined by the user-specified library. In general, for library-based methods, how to design an inclusive but not unnecessarily large library \emph{a priori} is important yet very challenging, which may require prior knowledge of the system to be identified and thus limit their applications for systems involving complex governing physics. {Moreover, in this work, only a uniform displacement of knots is used for spline representation, and the tensor product of 1-D splines is adopted for multi-dimensional spline constructions as shown in Tab.~\ref{tab:tp_spline}. These choices are not optimal and have notable limitations for high-dimensional problems.
To tackle this issue, we propose to apply more advanced spline learning techniques to reduce the computational cost and improve the scalability. For instance, using deep learning to optimize knot size has shown great promise for lowering computational costs in high-dimensional settings~\cite{laube2018deep,durkan2019neural}, while defining continuous spline kernels and expanding basis in subdomains will allow a significant reduction of trainable parameters~\cite{fey2018splinecnn}. The improvement of spline learning will be explored in our future work.} {Lastly, similarly to all data-driven models, the proposed method could have a negative societal impact if it is used abusively, particularly for predictive modeling of high-consequence natural systems, where caution should be taken for decision making.}




\begin{ack}
The authors would like to acknowledge the funds from National Science Foundation, United States of America under award numbers CMMI-1934300 and OAC-2047127, the Air Force Office of Scientific Research (AFOSR), United States of America under award number FA9550-22-1-0065, and startup funds from the College of Engineering at University of Notre Dame in supporting this study


\end{ack}






\bibliographystyle{elsarticle-num}
\bibliography{ref,ref_self}

\begin{thebibliography}{10}
\expandafter\ifx\csname url\endcsname\relax
  \def\url#1{\texttt{#1}}\fi
\expandafter\ifx\csname urlprefix\endcsname\relax\def\urlprefix{URL }\fi
\expandafter\ifx\csname href\endcsname\relax
  \def\href#1#2{#2} \def\path#1{#1}\fi

\bibitem{pfaff2020learning}
T.~Pfaff, M.~Fortunato, A.~Sanchez-Gonzalez, P.~Battaglia, Learning mesh-based
  simulation with graph networks, in: International Conference on Learning
  Representations, 2020.

\bibitem{lu2021learning}
L.~Lu, P.~Jin, G.~Pang, Z.~Zhang, G.~E. Karniadakis, Learning nonlinear
  operators via {D}eep{ON}et based on the universal approximation theorem of
  operators, Nature Machine Intelligence 3~(3) (2021) 218--229.

\bibitem{han2022predicting}
X.~Han, H.~Gao, T.~Pfaff, J.-X. Wang, L.~Liu, Predicting physics in
  mesh-reduced space with temporal attention, in: International Conference on
  Learning Representations, 2022.

\bibitem{brunton2016discovering}
S.~L. Brunton, J.~L. Proctor, J.~N. Kutz, Discovering governing equations from
  data by sparse identification of nonlinear dynamical systems, Proceedings of
  the national academy of sciences 113~(15) (2016) 3932--3937.

\bibitem{rudy2017data}
S.~H. Rudy, S.~L. Brunton, J.~L. Proctor, J.~N. Kutz, Data-driven discovery of
  partial differential equations, Science advances 3~(4) (2017) e1602614.

\bibitem{chu2020discovering}
H.~K. Chu, M.~Hayashibe, Discovering interpretable dynamics by sparsity
  promotion on energy and the lagrangian, IEEE Robotics and Automation Letters
  5~(2) (2020) 2154--2160.

\bibitem{rudy2019deep}
S.~H. Rudy, J.~N. Kutz, S.~L. Brunton, Deep learning of dynamics and
  signal-noise decomposition with time-stepping constraints, Journal of
  Computational Physics 396 (2019) 483--506.

\bibitem{wang2020denoising}
J.~Wang, X.~Xie, J.~Shi, W.~He, Q.~Chen, L.~Chen, W.~Gu, T.~Zhou, Denoising
  autoencoder, a deep learning algorithm, aids the identification of a novel
  molecular signature of lung adenocarcinoma, Genomics, proteomics \&
  bioinformatics 18~(4) (2020) 468--480.

\bibitem{wu2021semi}
H.~Wu, P.~Du, R.~Kokate, J.-X. Wang, A semi-analytical solution and {AI}-based
  reconstruction algorithms for magnetic particle tracking, Plos one 16~(7)
  (2021) e0254051.

\bibitem{long2019pde}
Z.~Long, Y.~Lu, B.~Dong, {PDE-Net} 2.0: Learning {PDE}s from data with a
  numeric-symbolic hybrid deep network, Journal of Computational Physics 399
  (2019) 108925.

\bibitem{kim2020integration}
S.~Kim, P.~Y. Lu, S.~Mukherjee, M.~Gilbert, L.~Jing, V.~{\v{C}}eperi{\'c},
  M.~Solja{\v{c}}i{\'c}, Integration of neural network-based symbolic
  regression in deep learning for scientific discovery, IEEE Transactions on
  Neural Networks and Learning Systems 32~(9) (2020) 4166--4177.

\bibitem{corbetta2020application}
M.~Corbetta, Application of sparse identification of nonlinear dynamics for
  physics-informed learning, in: 2020 IEEE Aerospace Conference, IEEE, 2020,
  pp. 1--8.

\bibitem{chen2021physics}
Z.~Chen, Y.~Liu, H.~Sun, Physics-informed learning of governing equations from
  scarce data, Nature Communications 12 (2021) 6136.

\bibitem{sun2021physics}
F.~Sun, Y.~Liu, H.~Sun, Physics-informed spline learning for nonlinear dynamics
  discovery, in: Proceedings of the Thirtieth International Joint Conference on
  Artificial Intelligence ({IJCAI-21}), 2021, pp. 2054--2061.

\bibitem{zhang2018robust}
S.~Zhang, G.~Lin, Robust data-driven discovery of governing physical laws with
  error bars, Proceedings of the Royal Society A: Mathematical, Physical and
  Engineering Sciences 474~(2217) (2018) 20180305.

\bibitem{zhang2019robust}
S.~Zhang, G.~Lin, Robust data-driven discovery of governing physical laws using
  a new subsampling-based sparse {Bayesian} method to tackle four challenges
  (large noise, outliers, data integration, and extrapolation), arXiv preprint
  arXiv:1907.07788 (2019).

\bibitem{hirsh2021sparsifying}
S.~M. Hirsh, D.~A. Barajas-Solano, J.~N. Kutz, Sparsifying priors for
  {Bayesian} uncertainty quantification in model discovery, arXiv preprint
  arXiv:2107.02107 (2021).

\bibitem{tipping2001sparse}
M.~E. Tipping, Sparse {Bayesian} learning and the relevance vector machine,
  Journal of machine learning research 1~(Jun) (2001) 211--244.

\bibitem{tipping2003fast}
M.~E. Tipping, A.~C. Faul, et~al., Fast marginal likelihood maximisation for
  sparse {Bayesian} models., in: AISTATS, 2003.

\bibitem{bishop2013variational}
C.~M. Bishop, M.~Tipping, Variational relevance vector machines, arXiv preprint
  arXiv:1301.3838 (2013).

\bibitem{faul2001variational}
A.~C. Faul, M.~E. Tipping, A variational approach to robust regression, in:
  International Conference on Artificial Neural Networks, Springer, 2001, pp.
  95--102.

\bibitem{faul2002analysis}
A.~C. Faul, M.~E. Tipping, Analysis of sparse {Bayesian} learning, in: Advances
  in neural information processing systems, 2002, pp. 383--389.

\bibitem{maddox2019simple}
W.~J. Maddox, P.~Izmailov, T.~Garipov, D.~P. Vetrov, A.~G. Wilson, A simple
  baseline for {Bayesian} uncertainty in deep learning, Advances in Neural
  Information Processing Systems 32 (2019).

\bibitem{quade2018sparse}
M.~Quade, M.~Abel, J.~Nathan~Kutz, S.~L. Brunton, Sparse identification of
  nonlinear dynamics for rapid model recovery, Chaos: An Interdisciplinary
  Journal of Nonlinear Science 28~(6) (2018) 063116.

\bibitem{champion2019data}
K.~Champion, B.~Lusch, J.~N. Kutz, S.~L. Brunton, Data-driven discovery of
  coordinates and governing equations, Proceedings of the National Academy of
  Sciences 116~(45) (2019) 22445--22451.

\bibitem{champion2019discovery}
K.~P. Champion, S.~L. Brunton, J.~N. Kutz, Discovery of nonlinear multiscale
  systems: Sampling strategies and embeddings, SIAM Journal on Applied
  Dynamical Systems 18~(1) (2019) 312--333.

\bibitem{zhang2019convergence}
L.~Zhang, H.~Schaeffer, On the convergence of the sindy algorithm, Multiscale
  Modeling \& Simulation 17~(3) (2019) 948--972.

\bibitem{tancik2020fourier}
M.~Tancik, P.~Srinivasan, B.~Mildenhall, S.~Fridovich-Keil, N.~Raghavan,
  U.~Singhal, R.~Ramamoorthi, J.~Barron, R.~Ng, Fourier features let networks
  learn high frequency functions in low dimensional domains, Advances in Neural
  Information Processing Systems 33 (2020) 7537--7547.

\bibitem{raissi2019physics}
M.~Raissi, P.~Perdikaris, G.~E. Karniadakis, Physics-informed neural networks:
  A deep learning framework for solving forward and inverse problems involving
  nonlinear partial differential equations, Journal of Computational Physics
  378 (2019) 686--707.

\bibitem{both2021deepmod}
G.-J. Both, S.~Choudhury, P.~Sens, R.~Kusters, Deepmod: Deep learning for model
  discovery in noisy data, Journal of Computational Physics 428 (2021) 109985.

\bibitem{wandel2021spline}
N.~Wandel, M.~Weinmann, M.~Neidlin, R.~Klein, Spline-pinn: Approaching pdes
  without data using fast, physics-informed hermite-spline cnns, arXiv preprint
  arXiv:2109.07143 (2021).

\bibitem{blundell2015weight}
C.~Blundell, J.~Cornebise, K.~Kavukcuoglu, D.~Wierstra, Weight uncertainty in
  neural network, in: International conference on machine learning, PMLR, 2015,
  pp. 1613--1622.

\bibitem{liu2016stein}
Q.~Liu, D.~Wang, Stein variational gradient descent: A general purpose
  {Bayesian} inference algorithm, Advances in neural information processing
  systems 29 (2016).

\bibitem{blei2017variational}
D.~M. Blei, A.~Kucukelbir, J.~D. McAuliffe, Variational inference: A review for
  statisticians, Journal of the American statistical Association 112~(518)
  (2017) 859--877.

\bibitem{gal2016dropout}
Y.~Gal, Z.~Ghahramani, Dropout as a {Bayesian} approximation: Representing
  model uncertainty in deep learning, in: international conference on machine
  learning, PMLR, 2016, pp. 1050--1059.

\bibitem{Ebrahimi2020Uncertainty-guided}
S.~Ebrahimi, M.~Elhoseiny, T.~Darrell, M.~Rohrbach, Uncertainty-guided
  continual learning with bayesian neural networks, in: International
  Conference on Learning Representations, 2020.

\bibitem{ritter2018scalable}
H.~Ritter, A.~Botev, D.~Barber, A scalable laplace approximation for neural
  networks, in: 6th International Conference on Learning Representations, ICLR
  2018-Conference Track Proceedings, Vol.~6, International Conference on
  Representation Learning, 2018.

\bibitem{wilson2020bayesian}
A.~G. Wilson, P.~Izmailov, {Bayesian} deep learning and a probabilistic
  perspective of generalization, Advances in neural information processing
  systems 33 (2020) 4697--4708.

\bibitem{abdar2021review}
M.~Abdar, F.~Pourpanah, S.~Hussain, D.~Rezazadegan, L.~Liu, M.~Ghavamzadeh,
  P.~Fieguth, X.~Cao, A.~Khosravi, U.~R. Acharya, et~al., A review of
  uncertainty quantification in deep learning: Techniques, applications and
  challenges, Information Fusion 76 (2021) 243--297.

\bibitem{frerix2021variational}
T.~Frerix, D.~Kochkov, J.~Smith, D.~Cremers, M.~Brenner, S.~Hoyer, Variational
  data assimilation with a learned inverse observation operator, in:
  International Conference on Machine Learning, PMLR, 2021, pp. 3449--3458.

\bibitem{gronquist2021deep}
P.~Gr{\"o}nquist, C.~Yao, T.~Ben-Nun, N.~Dryden, P.~Dueben, S.~Li, T.~Hoefler,
  Deep learning for post-processing ensemble weather forecasts, Philosophical
  Transactions of the Royal Society A 379~(2194) (2021) 20200092.

\bibitem{mccabe2021learning}
J.~McCabe, J.~Brown, Learning to assimilate in chaotic dynamical systems,
  Advances in Neural Information Processing Systems 34 (2021) 12237--12250.

\bibitem{fey2018splinecnn}
M.~Fey, J.~E. Lenssen, F.~Weichert, H.~M{\"u}ller, Splinecnn: Fast geometric
  deep learning with continuous {B}-spline kernels, in: Proceedings of the IEEE
  Conference on Computer Vision and Pattern Recognition, 2018, pp. 869--877.

\bibitem{izmailov2020subspace}
P.~Izmailov, W.~J. Maddox, P.~Kirichenko, T.~Garipov, D.~Vetrov, A.~G. Wilson,
  Subspace inference for {Bayesian} deep learning, in: Uncertainty in
  Artificial Intelligence, PMLR, 2020, pp. 1169--1179.

\bibitem{tipping1999relevance}
M.~Tipping, The relevance vector machine, Advances in neural information
  processing systems 12 (1999).

\bibitem{wang2021long}
S.~Wang, P.~Perdikaris, Long-time integration of parametric evolution equations
  with physics-informed deeponets, arXiv preprint arXiv:2106.05384 (2021).

\bibitem{laube2018deep}
P.~Laube, M.~O. Franz, G.~Umlauf, Deep learning parametrization for {B}-spline
  curve approximation, in: 2018 International Conference on 3D Vision (3DV),
  IEEE, 2018, pp. 691--699.

\bibitem{durkan2019neural}
C.~Durkan, A.~Bekasov, I.~Murray, G.~Papamakarios, Neural spline flows,
  Advances in neural information processing systems 32 (2019).

\bibitem{de1968convergence}
C.~de~Boor, On the convergence of odd-degree spline interpolation, Journal of
  approximation theory 1~(4) (1968) 452--463.

\bibitem{swartz1968O}
B.~Swartz, ${O}(h^{2n+2-l})$ bounds on some spline interpolation errors,
  Bulletin of the American Mathematical Society 74~(6) (1968) 1072--1078.

\bibitem{huang2004polynomial}
J.~Z. Huang, C.~O. Wu, L.~Zhou, Polynomial spline estimation and inference for
  varying coefficient models with longitudinal data, Statistica Sinica (2004)
  763--788.

\bibitem{de1978practical}
C.~De~Boor, C.~De~Boor, A practical guide to splines, Vol.~27, springer-verlag
  New York, 1978.

\end{thebibliography}

\clearpage
\section*{Checklist}

Please do not modify the questions and only use the provided macros for your
answers.  Note that the Checklist section does not count towards the page
limit.  In your paper, please delete this instructions block and only keep the
Checklist section heading above along with the questions/answers below.

\begin{enumerate}

\item For all authors...
\begin{enumerate}
  \item Do the main claims made in the abstract and introduction accurately reflect the paper's contributions and scope?
    \answerYes{}
  \item Did you describe the limitations of your work?
    \answerYes{See Section~\ref{sec:conclusion}}
  \item Did you discuss any potential negative societal impacts of your work?
    \answerYes{}
  \item Have you read the ethics review guidelines and ensured that your paper conforms to them?
    \answerYes{}
\end{enumerate}

\item If you are including theoretical results...
\begin{enumerate}
  \item Did you state the full set of assumptions of all theoretical results?
    \answerYes{See Section~\ref{sec:meth}}
        \item Did you include complete proofs of all theoretical results?
    \answerYes{See supplemental materials}
\end{enumerate}

\item If you ran experiments...
\begin{enumerate}
  \item Did you include the code, data, and instructions needed to reproduce the main experimental results (either in the supplemental material or as a URL)?
    \answerYes{It is included as a URL}
  \item Did you specify all the training details (e.g., data splits, hyperparameters, how they were chosen)?
    \answerYes{See supplemental materials}
        \item Did you report error bars (e.g., with respect to the random seed after running experiments multiple times)?
    \answerYes{See Section~\ref{sec:result} and supplemental materials}
        \item Did you include the total amount of compute and the type of resources used (e.g., type of GPUs, internal cluster, or cloud provider)?
    \answerYes{See supplemental materials}
\end{enumerate}

\item If you are using existing assets (e.g., code, data, models) or curating/releasing new assets...
\begin{enumerate}
  \item If your work uses existing assets, did you cite the creators?
    \answerYes{}
  \item Did you mention the license of the assets?
    \answerNo{It is general open source code available in public for research purpose}
  \item Did you include any new assets either in the supplemental material or as a URL?
    \answerYes{}
  \item Did you discuss whether and how consent was obtained from people whose data you're using/curating?
    \answerNA{}
  \item Did you discuss whether the data you are using/curating contains personally identifiable information or offensive content?
    \answerNA{}
\end{enumerate}

\item If you used crowdsourcing or conducted research with human subjects...
\begin{enumerate}
  \item Did you include the full text of instructions given to participants and screenshots, if applicable?
    \answerNA{}
  \item Did you describe any potential participant risks, with links to Institutional Review Board (IRB) approvals, if applicable?
    \answerNA{}
  \item Did you include the estimated hourly wage paid to participants and the total amount spent on participant compensation?
    \answerNA{}
\end{enumerate}

\end{enumerate}


\clearpage
\appendix

\section{Appendix}
\label{sec:app}
\subsection{Proof: analytical form of spline derivatives}
\label{subsec:spderiv}
\begin{theorem}. The derivative of B-spline curve can be analytically evaluated as:
\begin{equation}
    \dot{y}(t)=\Sigma_{s=0}^{r+k-1}\dot{N}_{s,k}(t)\theta_s,
\end{equation}
where 
\begin{equation}
\label{eq:appSplineDeriv}
    \dot{N}_{s,k}(t) =\frac{k}{\tau_{s+k}-\tau_{s}}N_{s,k-1}(t)-\frac{k}{\tau_{s+k+1}-\tau_{s+1}}N_{s+1,k-1}(t).
\end{equation}
\end{theorem}
\begin{proof}
We will prove by the induction. 
For the base case $k=1$, we have 
\begin{equation}
    \dot{N}_{s,1}(t) =
    \begin{cases}
    \frac{1}{\tau_{s+1}-\tau_{s}} & t\in [\tau_{s}, \tau_{s+1}] \\
    -\frac{1}{\tau_{s+2}-\tau_{s+1}} & t\in [\tau_{s+1}, \tau_{s+2}] 
    \end{cases}
    =
    \frac{1}{\tau_{s+1}-\tau_{s}}N_{s,0}(t)-\frac{1}{\tau_{s+2}-\tau_{s+1}}N_{s+1,0}(t).
\end{equation}
Let's suppose that the formula holds for $k$ up to $n$. We will prove that this formula also holds for $k = n+1$. By the definition in Eq.~\ref{eq:spline1D} and the chain rule, we can get that:
\begin{equation}
\begin{aligned}
\dot{N}_{s,n+1}(t) 
&=\frac{t-\tau_s}{\tau_{s+n+1}-\tau_{s}}\dot{N}_{s,n}(t)+\frac{N_{s,n}(t)}{\tau_{s+n+1}-\tau_{s}}+ \frac{\tau_{s+n+2}-t}{\tau_{s+n+2}-\tau_{s+1}}\dot{N}_{s+1,n}(t)-\frac{N_{s+1,n}(t)}{\tau_{s+n+2}-\tau_{s+1}}\\
&= \frac{t-\tau_s}{\tau_{s+n+1}-\tau_{s}}\bigg(\frac{n}{\tau_{s+n}-\tau_{s}}N_{s,n-1}(t) - \frac{n}{\tau_{s+n+1}-\tau_{s+1}}N_{s+1,n-1}(t)\bigg) \\
&+\frac{\tau_{s+n+2}-t}{\tau_{s+n+2}-\tau_{s+1}}\bigg(\frac{n}{\tau_{s+n+1}-\tau_{s+1}}N_{s+1,n-1}(t) - \frac{n}{\tau_{s+n+2}-\tau_{s+2}}N_{s+2,n-1}(t)\bigg)\\
&+ \frac{1}{\tau_{s+n+1}-\tau_{s}}N_{s,n}(t)-\frac{1}{\tau_{s+n+2}-\tau_{s+1}}N_{s+1,n}(t)\\
& =\frac{n+1}{\tau_{s+n+1}-\tau_{s}}N_{s,n}(t)-\frac{n+1}{\tau_{s+n+2}-\tau_{s+1}}N_{s+1,n}(t).
\end{aligned}
\end{equation}

\end{proof}

\subsection{Spline representation}
\label{subsec:sprep}

In this section, we give error bounds for spline representation. 
For simplicity, we consider 1D scenario and assume the target function $u:[0,1]\rightarrow R$ is periodic and defined on the unit interval $\Omega = [0,1]$. 
Consider a set of uniform knots $\Gamma: 0=\tau_0 \leq \tau_1\leq \cdots \leq \tau_{r+k} = 1$ with .
The space of $k$th degree $\{N_{s,k}\}_{s=0}^{r+k-1}$ splines is
\begin{align*}
S^{k}(\Omega, \Gamma) = \{p|p(t) \textrm{ is a polynomial of degree $k$ in each } (\tau_i, \tau_{i+1})\} \cap \mathcal{C}^{k-1}(\Omega).
\end{align*}
Spline interpolation seeks $\pred{u} \in S^{k}(\Omega, \Gamma)$ that satisfies $u(\tau_i) = \pred{u}(\tau_i) \quad \forall \quad 0 \leq r+k$.

\begin{theorem}[Spline interpolation error bounds~\cite{de1968convergence,swartz1968O}]
Assume that $u$ is periodic, for odd degree spline interpolation $k$, we have 
\begin{equation}
\label{theorem:fitting error}
\begin{aligned}
&\lVert u - \pred{u} \rVert_{\infty} =  \Delta \tau^{k+1}\Big(C_{k+1}\lVert u^{(k+1)}\rVert_{\infty} + \mathcal{O}(\omega(u^{(k+1)}, \Delta \tau)) \Big),\\
&\lVert u^{(l)} - \pred{u}^{(l)} \rVert_{\infty} =  \Delta \tau^{k+1-l}\Big(D_{k+1-l}\lVert u^{(k+1)}\rVert_{\infty} + \mathcal{O}(\omega(u^{(k+1)}, \Delta \tau)) \Big) \quad \forall 1\leq l \leq k-1.
\end{aligned}
\end{equation}
Here $C_{k+1}$ and $D_{k+1-l}$ are constant parameters, which are independent of $\Delta \tau$ and $u$, and $\omega(u^{(k+1)}, \Delta \tau) = \sup_{|x-y|\leq \Delta \tau}|u^{(k+1)}(x) - u^{(k+1)}(y)|$.
\end{theorem}

In the present work, we focus on using spline for smoothing noisy data. It is essentially a nonparameteric estimation of function $u$ from noisy data $\{t_i, \obs{u}_i\}_{i=1}^{n}$,
\begin{align*}
\obs{u}_i = u(t_i) + \eta_i \quad \textrm{ with } \quad \eta_i \sim \mathcal{N}(0,\delta_{\eta}^2).
\end{align*}
We further assume that $\{t_i\}$ are uniformly sampled from $\Omega$.
The unregularized smoothing process estimates spline basis coefficients 
$$\pred{\boldsymbol{\theta}} = (\Nm^T \Nm)^{-1} \Nm^T \cdot \obs{\mathbf{U}} $$
through minimizing $\lVert \obs{\mathbf{U}} - \Nm \cdot \boldsymbol{\theta}\rVert$. Here $\obs{\mathbf{U}} = \bigl[\obs{u}_1, \obs{u}_2, \cdots , \obs{u}_n\bigr]^T$ and $\Nm$ is the spline basis matrix evaluated at these measurement locations.
The optimal $\boldsymbol{\theta}^{\rm opt}$ satisfies $\pred{u}(t) = \mathbf{N}(t) \cdot \boldsymbol{\theta}^{\rm opt}$, where $\pred{u}$ is the spline interpolation of $u$, and $\mathbf{N}(t) = [N_{0,k}(t),N_{1,k}(t),\cdots,N_{r+k-1,k}(t)]^T$ denotes spline basis function vector.
Following~\cite{huang2004polynomial}, we have spline fitting error bounds, as following.

\begin{theorem}[Spline fitting error bounds] Assume that $u$ is periodic and the number of data $n$ is sufficient large, for odd degree spline interpolation $k$, we have
\begin{align*} 
   &\lVert  \mathbb{E} \pred{\boldsymbol{\theta}} - \boldsymbol{\theta}^{\rm opt} \lVert_{2} \leq C_1 \Delta \tau^{k+1}  \qquad \lVert  \mathrm{Cov}\pred{\boldsymbol{\theta}} \lVert_{2} \leq C_2 \frac{\delta_\eta^2}{n},   
\end{align*}
where $C_1$ and $C_2$ are constant and independent of $n$.
\end{theorem}

\begin{proof}
Let us denote 
\begin{align*}
    &\mathbf{U} = [u(t_1),u(t_2),\cdots,u(t_n)]^T \quad
    \pred{\mathbf{U}} = [\pred{u}(t_1),\pred{u}(t_2),\cdots,\pred{u}(t_n)]^T \\
    &\boldsymbol{\eta} = [\eta_1,\eta_2,\cdots,\eta_n]^T  \quad
    \mathbf{e} = \mathbf{U} - \pred{\mathbf{U}}.
\end{align*}
We have
\begin{equation}
\label{eq:theta-diff}
\begin{split}
\pred{\boldsymbol{\theta}} - \boldsymbol{\theta}^{opt} 
    &= (\Nm^T \Nm)^{-1} \Nm^T \cdot \obs{\mathbf{U}} - (\Nm^T \Nm)^{-1} \Nm^T\Nm {\boldsymbol{\theta}}^{\rm opt}  \\
    &= (\Nm^T \Nm)^{-1} \Nm^T \cdot (\mathbf{U} + \boldsymbol{\eta} - \pred{\mathbf{U}} ) \\
    &= (\Nm^T \Nm)^{-1} \Nm^T \cdot (\mathbf{e} + \boldsymbol{\eta}).
\end{split}
\end{equation}
We will first prove that
\begin{align}
\label{eq:matrix-norm}
\lVert \Nm^T \Nm \rVert_2  = \mathcal{O}(n) \qquad \lVert (\Nm^T \Nm)^{-1} \rVert_2  = \mathcal{O}(\frac{1}{n}).
\end{align}
For any $\boldsymbol{\theta}$, $\frac{1}{n}\boldsymbol{\theta}^T \Nm^T \Nm\boldsymbol{\theta}$ is the Monte Carlo approximation of $\int \bigl(\mathbf{N}(t)^T \cdot \boldsymbol{\theta} \bigr)^2 dt $, and hence 

\begin{align}
\label{eq:MCMC-S}
\frac{1}{n}\boldsymbol{\theta}^T\Nm^T \Nm \boldsymbol{\theta} = \int (\mathbf{N}(t)^T  \boldsymbol{\theta})^2 dt   + \mathcal{O}(\frac{1}{\sqrt{n}}).
\end{align}

Bringing the following property of B-splines~\cite{de1978practical}
\begin{align*}
M_1 \boldsymbol{\theta}^T\boldsymbol{\theta} \leq \int (\mathbf{N}(t)^T  \boldsymbol{\theta})^2 dt \leq M_2 \boldsymbol{\theta}^T\boldsymbol{\theta} \qquad \exists M_1, M_2 > 0
\end{align*}
into Eq.~\eqref{eq:MCMC-S} leads to Eq.~\eqref{eq:matrix-norm}. Then we prove that 
\begin{align}
\label{eq:error}
\lVert \Nm^T \mathbf{e} \rVert_2 = \mathcal{O}(n \Delta \tau^{k+1}).
\end{align}
Since $\sum_s N_{s,k}(t) = 1$ and $\lVert \mathbf{e}\rVert_\infty = \mathcal{O}(\Delta \tau^{k+1})$, we have
\begin{align*}
\lVert \Nm^T \mathbf{e} \rVert_2 \leq \lVert \Nm^T \mathbf{e} \rVert_1 = \mathcal{O}(n \Delta \tau^{k+1}).
\end{align*}

Finally, combining Eq.~\eqref{eq:theta-diff}, Eq.~\eqref{eq:matrix-norm} and Eq.~\eqref{eq:error} leads to  
\begin{align*} 
   &\lVert  \mathbb{E} \pred{\boldsymbol{\theta}} - \boldsymbol{\theta}^{\rm opt} \lVert_{2} = \lVert (\Nm^T \Nm)^{-1} \Nm^T \cdot \mathbf{e}\lVert_{2} \leq \lVert (\Nm^T \Nm)^{-1} \lVert_{2} \lVert \Nm^T \cdot \mathbf{e}\lVert_{2} \leq C_1 \Delta \tau^{k+1}\\
   &\lVert  \mathrm{Cov}\pred{\boldsymbol{\theta}} \lVert_{2} = \sigma_{\eta}^2\lVert  (\Nm^T \Nm)^{-1} \lVert_{2} \leq C_2\frac{\delta_\eta^2}{n}
\end{align*}
\end{proof}

In our sparse Bayesian regression, in stead of solving the aforementioned minimization problem, we have additional regularization terms.

\subsection{Algorithms}
\label{subsec:algs}
In this section, we present detailed algorithms used in the present work, which include Bayesian Alternative Direction Optimization~(ADO) Learning~\ref{alg1},  Sequential Threshold Sparse Bayesian Learning~\ref{alg2}, and Ensemble Kalman Filter~\ref{alg3}.

\begin{algorithm}[h!]
\caption{Bayesian Alternative Direction Optimization~(ADO) Learning}\label{alg1}
    \footnotesize
    \SetAlgoLined
    \SetKwInOut{Input}{Input}
    \SetKwInOut{Output}{Output}
    \Input{Library $\mathbf{\Phi}$, spline basis $\mathbf{N}$, time derivative of spline basis matrix $\dNc$, negative log form of equation Eq.~\ref{eq:posterior} $\mathcal{L}$}
    \Output{Mean estimation: $\boldsymbol{\theta}_{\text{SWA}}$, $\mathbf{W}_{\text{SWA}}$, $\mathbf{B}_{\text{SWA}}$, $\mathbf{P}_{\text{SWA}}$ \\
    Samples from posterior distributions: $\sample{\boldsymbol{\theta}}$, $\sample{\mathbf{W}}$, $\sample{\mathbf{B}}$, $\sample{\mathbf{P}}$ \\
    }
    \textbf{Pretrain}:\\
    \For{$i = 1 : T_{\rm Pretrain}$}{
        \textbf{SDG optimization with the fixed library}\\
        $\{\boldsymbol{\theta}_{i+1}, \mathbf{W}'_{i+1}, \mathbf{B}_{i+1}, \mathbf{P}'_{i+1}\} = \argmin{\mathcal{L}}$\\
        \textbf{Update library}\\
        $\mathbf{W}_{i+1},\mathbf{P}_{i+1} = \text{STSparseBayesian}(\mathbf{\Phi},\dot{\mathbf{U}}_{i+1}=\dNc\boldsymbol{\theta}_{i+1}, \mathbf{W}'_{i+1}, \mathbf{P}'_{i+1})$\\
        \uIf{$\mathcal{L}(\boldsymbol{\theta}_{i+1},\mathbf{W}_{i+1},\mathbf{B}_{i+1},  \mathbf{P}_{i+1}) < \mathcal{L}^{\star}$}{
        $\mathcal{L}^{\star} =\mathcal{L}(\boldsymbol{\theta}_{i+1},\mathbf{W}_{i+1}, \mathbf{B}_{i+1}, \mathbf{P}_{i+1}) $\\
        $ \boldsymbol{\theta}^{\star},\mathbf{W}^{\star}, \mathbf{B}^{\star}, \mathbf{P}^{\star}= \boldsymbol{\theta}_{i+1},\mathbf{W}_{i+1}, \mathbf{B}_{i+1}, \mathbf{P}_{i+1}$
        }
        \uElse{break}
    }
    \textbf{Stochastic Weight Averaging-Gaussian~(SWAG) for posterior approximation}:\\
    $\boldsymbol{\theta}_{\text{SWA}}, \mathbf{W}_{\text{SWA}},  \mathbf{B}_{\text{SWA}}, \mathbf{P}_{\text{SWA}}= \boldsymbol{\theta^{\star}},\mathbf{W}^{\star},  \mathbf{B}^{\star}, \mathbf{P}^{\star}$\\
    \text{With a constant learning rate} \For{$i = 1 : T_{\rm SWAG}$}{
        SGD update $\boldsymbol{\theta}_{i},\mathbf{W}_{i}, \mathbf{B}_{i}, \mathbf{P}_{i}$\\
        $\boldsymbol{\theta}_{\text{SWA}}, \mathbf{W}_{\text{SWA}},  \mathbf{B}_{\text{SWA}}, \mathbf{P}_{\text{SWA}}=\frac{i\boldsymbol{\theta}_{\text{SWA}}+\boldsymbol{\theta}_i}{i+1}, \frac{i\boldsymbol{W}_{\text{SWA}}+\boldsymbol{W}_i}{i+1},\frac{i\boldsymbol{B}_{\text{SWA}}+\boldsymbol{B}_i}{i+1}, \frac{i\boldsymbol{P}_{\text{SWA}}+\boldsymbol{P}_i}{i+1}$
    }
    \textbf{Compute low-rank square root of empirical covariance matrices}
    $\mathbf{\Lambda}_{\theta}, \mathbf{\Lambda}_{W}, \mathbf{\Lambda}_{B}, \mathbf{\Lambda}_{P}$ from 
    $\{\boldsymbol{\theta}_{i}-\boldsymbol{\theta}_{\text{SWA}}\},\{\boldsymbol{W}_{i}-\boldsymbol{W}_{\text{SWA}}\}, \{\boldsymbol{B}_{i}-\boldsymbol{B}_{\text{SWA}}\}, \{\boldsymbol{P}_{i}-\boldsymbol{P}_{\text{SWA}}\}$\\
    \textbf{Sampling}:\\
    $\sample{\boldsymbol{\theta}}= \boldsymbol{\theta}_{\text{SWA}}+\mathbf{\Lambda}_{\theta}\sample{\boldsymbol{z}}_\theta \qquad \sample{\boldsymbol{W}}=\boldsymbol{W}_{\text{SWA}}+\mathbf{\Lambda}_{W}\sample{\boldsymbol{z}}_W \qquad \sample{\boldsymbol{B}}=\boldsymbol{B}_{\text{SWA}}+\mathbf{\Lambda}_{B}\sample{\boldsymbol{z}}_B \qquad \sample{\boldsymbol{P}}=\boldsymbol{P}_{\text{SWA}}+\mathbf{\Lambda}_{P}\sample{\boldsymbol{z}}_P$\\
    \text{where} $\sample{\boldsymbol{z}}_*$ \text{are the random samples from} $\mathcal{N}(0, I)$.
\end{algorithm}

\begin{algorithm}[h!]
	\caption{Sequential Threshold Sparse Bayesian Learning}\label{alg2}
	\footnotesize
	\SetAlgoLined
	\SetKwInOut{Input}{Input}
    \SetKwInOut{Output}{Output}
    \Input{Spline trainable parameter $\boldsymbol{\theta}$,  library $\mathbf{\Phi}(\boldsymbol{\theta})$, approximated derivative $\dot{\mathbf{U}}$, library weight $\mathbf{W}$, and process error matrix $\mathbf{P}$}
    \Output{Best solution library $\mathbf{\Phi}^{\star}$,$\mathbf{W}^{\star}$,  $\mathbf{P}^{\star}$}
	
	\SetKwInOut{Initialize}{Initialize}
	\Initialize {Threshold $\epsilon$, number of library terms $p_{\text{old}}$, and Flag = $\mathbf{True}$}
	\While{Flag is True}{
	\While{not converged}{
    	1. Compute the relevance variable $\eta_i = {q_i}^2 - s_i$ as defined in ~\cite{tipping2003fast}\\
        2. Update library $\boldsymbol{\Phi}^{\star}$, weight $\mathbf{W}^{\star}$, and process error matrix $\mathbf{P}^{\star}$, as shown in ~\cite{tipping2003fast}
    }
        
        \For{$j = 1 : p_{\text{old}}$}{
    	$\mathbf{W}^{\star}(j) = 0$ {$\textbf{If}$ $|\mathbf{W}^{\star}(j)|\leq \epsilon$
    	}}
    	Find the nonzero entries in $\mathbf{W}^{\star}$, record the index as $\mathbf{I}$,
    	update $\mathbf{\Phi} = \mathbf{\Phi}(:, \mathbf{I})$
         and $p_{\text{new}}$ = length of $\mathbf{I}$\;
         \uIf {$\mathbf{p_{new} = p_{old}}$}{Flag = $\mathbf{False}$}
     }
\end{algorithm}
\clearpage
\begin{algorithm}
	\caption{Ensemble Kalman Filter}\label{alg3}
	\footnotesize
	\SetAlgoLined
	\SetKwInOut{Input}{Input}
    \SetKwInOut{Output}{Output}
    \Input{ensemble number $J$, sampled weights $\{\mathbf{W}^{j}\}_{j=1}^{J}$, discovered dynamical model $\mathbf{M}(\bigcdot{}\ ;\ \space\mathbf{W})$, process noise covariance $\mathbf{P}$,  observation model $\mathbf{h}$,  observation $\obs{\mathbf{U}}$, observation noise covariance $\mathbf{B}$}
    \Output{Analysis ensemble trajectories $\mathbf{U}_a^{j}(t)$}

	\SetKwInOut{Forecast}{Forecast}
	\Forecast{$\mathbf{U}_f^{j}(t_{i+1})=\mathbf{M}(\mathbf{U}_a^{j}(t_{i});\mathbf{W}^{j})+\boldsymbol{\epsilon}^j_2,\quad\boldsymbol{\epsilon}_2^{j}\sim \mathcal (\mathbf{0}, \mathbf{P})$
	\\
	$\overline{\mathbf{U}}_f(t_{i+1}) = \frac{1}{J}\sum_{j=1}^{J}{\mathbf{U}_f^{j}(t_{i+1})}$\\
    }
	\SetKwInOut{Analysis}{Analysis}
	\Analysis{${\mathbf{U}}_h^{j}(t_{i+1}) = \mathbf{h}({\mathbf{U}}_f^{j}(t_{i+1}))
	\qquad 
	\overline{\mathbf{U}}_h(t_{i+1}) = \frac{1}{J}\sum_{j=1}^{J}{\mathbf{U}_h^{j}(t_{i+1})}$ 
	$\mathbf{C}^{fh}(t_{i+1})=\frac{1}{{J}-1}\Sigma_{j=1}^{J}(\mathbf{U}^i_f(t_{i+1}) - \overline{\mathbf{U}}_f(t_{i+1}))(\mathbf{U}^i_h(t_{i+1}) - \overline{\mathbf{U}}_h(t_{i+1}))^{T} $
	$\mathbf{C}^{hh}(t_{i+1})=\frac{1}{{J}-1}\Sigma_{j=1}^{J}(\mathbf{U}^i_h(t_{i+1}) - \overline{\mathbf{U}}_h(t_{i+1}))(\mathbf{U}^i_h(t_{i+1}) - \overline{\mathbf{U}}_h(t_{i+1}))^{T}  + \mathbf{B}$
	$\mathbf{K}(t_{i+1})= \mathbf{C}^{fh}\bigl(\mathbf{C}^{hh}\bigr)^{-1}$\\
	${\mathbf{U}}_a^{j}(t_{i+1}) = {\mathbf{U}}_f^{j}(t_{i+1})+\mathbf{K}(t_{i+1})\bigg(\obs{\mathbf{U}}(t_{i+1})-{\mathbf{U}}_h^{j}(t_{i+1})  - \boldsymbol{\epsilon}_1^{j}\bigg),\quad\boldsymbol{\epsilon}_1^{j}\sim \mathcal (\mathbf{0}, \mathbf{B})$}
\end{algorithm}

\subsection{Training Details}
Additional training hyper parameters used in Sec.~\ref{sec:result} is shown in the Tab.~\ref{tab:trainDetail}.
  \begin{table}[H]
  \caption{Training Details}
  \label{tab:trainDetail}
  \vspace{6pt}
  \centering
  \footnotesize
  \begin{tabular}{lllll}
    \toprule
    Case&Van der Pol& Lorenz 96&Advection&Burgers'
    \\
    \midrule
    ADO Iter & $5$  &$5$&$1$& $5$    \\
    ADO Epoch & $20K$  &$50K$&$20K$& $20K$    \\
    Post Epoch & $1K$  &$65K$&$2K$& $0.5K$    \\
    SWAG Epoch & $1.5K$  &$80K$&$0.5K$& $0.5K$ 
    \\
    LR&$1\times10^{-2}$&$1\times10^{-2}$&$1\times10^{-2}$&$1\times10^{-2}$\\
    SWAG LR&$1\times10^{-3}$&$1\times10^{-3}$&$1\times10^{-3}$&$1\times10^{-3}$\\

    \bottomrule
  \end{tabular}
\end{table}
\subsection{Additional Result: ODE}
\label{subsec:ODE1}
\begin{figure}[H]
   \centering
   \includegraphics[width=1.0\textwidth]{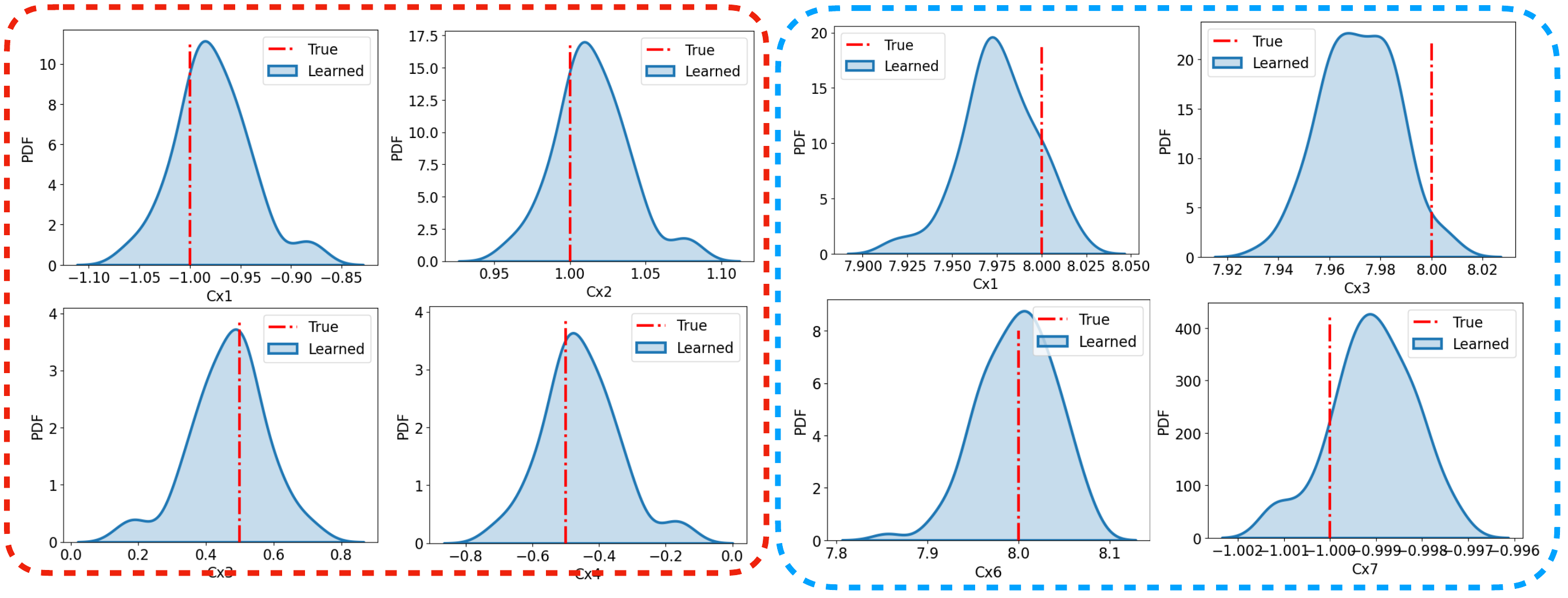}
   \caption{Additional discovery results for ODE systems; the red box shows results coefficients distribution for the Van der Pol system, and the blue box shows the result for Lorenz 96 system.}
   \label{fig:addODE:L96}
 \end{figure}

\begin{figure}[H]
   \centering
   \includegraphics[width=1.0\textwidth]{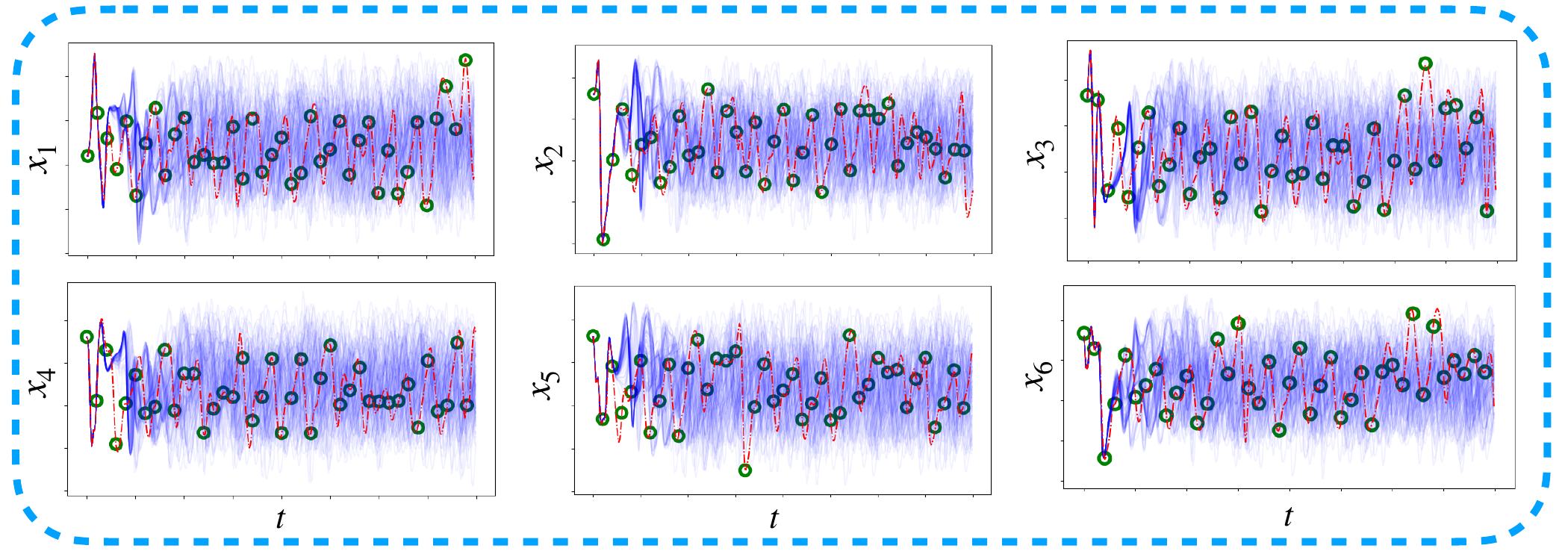}
   \caption{Additional UQ results for ODE systems; the blue box shows the all the states prediction for Lorenz 96 system without ensemble Kalman filter.}
   \label{fig:addODE}
 \end{figure}

\begin{figure}[H]
   \centering
   \includegraphics[width=1.0\textwidth]{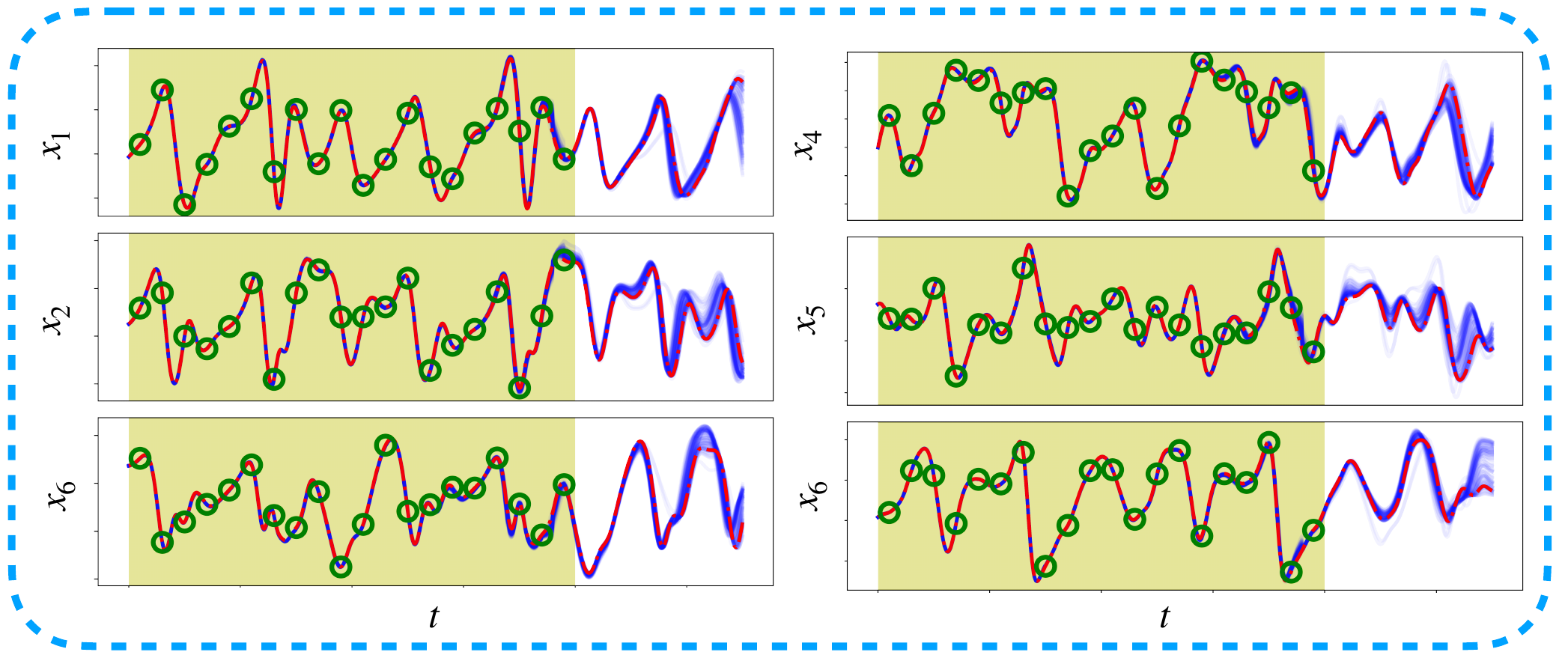}
   \caption{Additional UQ results for ODE systems; the blue box shows the all the states prediction for Lorenz 96 system with ensemble Kalman filter.}
   \label{fig:addUQODE}
 \end{figure}

We list additional discovery and UQ results in this section. Fig.~\ref{fig:addODE} shows 4 distributions of the coefficients for Van der Pol system in red box and Lorenz 96 system in blue box. {Fig.~\ref{fig:addODE:L96} shows additional UQ result from the identified L96 systems without incorporating the data assimilation process. The truth trajectory is marked by red. The measurement is marked by green dots and the ensemble trajectories are marked by blue. Although the system has been identified with high accuracy, as shown in Tab.~\ref{tab:FullODEcp1}, the predicted ensembles of the state variables still become chaotic after several seconds. It is inevitable since the chaotic nature of the underlying system, which means any small perturbation in any parameters would significantly influence the future trajectories. Fortunately, the predicted covariance matrix of the Bayesian framework makes it easy to incorporate the data assimilation with the identified systems. With the identified distribution of system coefficients, the data assimilation can be used to predict the future states with reduced uncertainty, given noisy measurement data in the past. Fig.~\ref{fig:addUQODE} shows additional UQ result for all the 6 state variables for Lorenz 96 system, incorporating EnKF algorithms.} 
 
\subsection{Additional Result: PDE}
\label{subsec:PDE1}

\begin{figure}[H]
  \centering
  \includegraphics[width=1.0\textwidth]{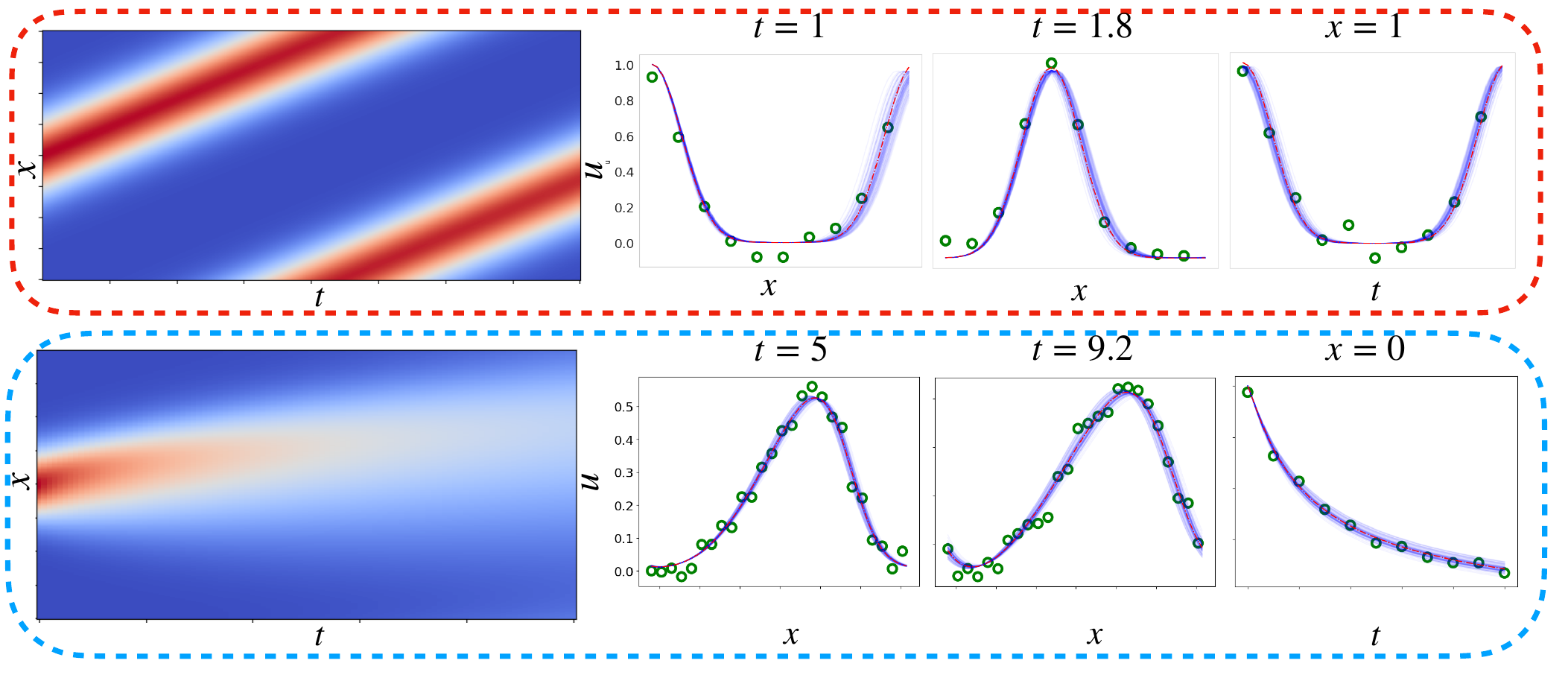}
  \caption{The discovery results for PDE systems; the red box shows results for the advection system, and the blue box shows the result for the Burgers' system. The layout inside each box follows the rules below. Leftmost sub-fig: true contour plot; Middle two sub-figs: the spatial results at different time $t$; Rightmost sub-figs: the temporal result at a fixed point $x$.}
  \label{fig:PDE}
 \end{figure}

\begin{figure}[H]
   \includegraphics[width=1\textwidth]{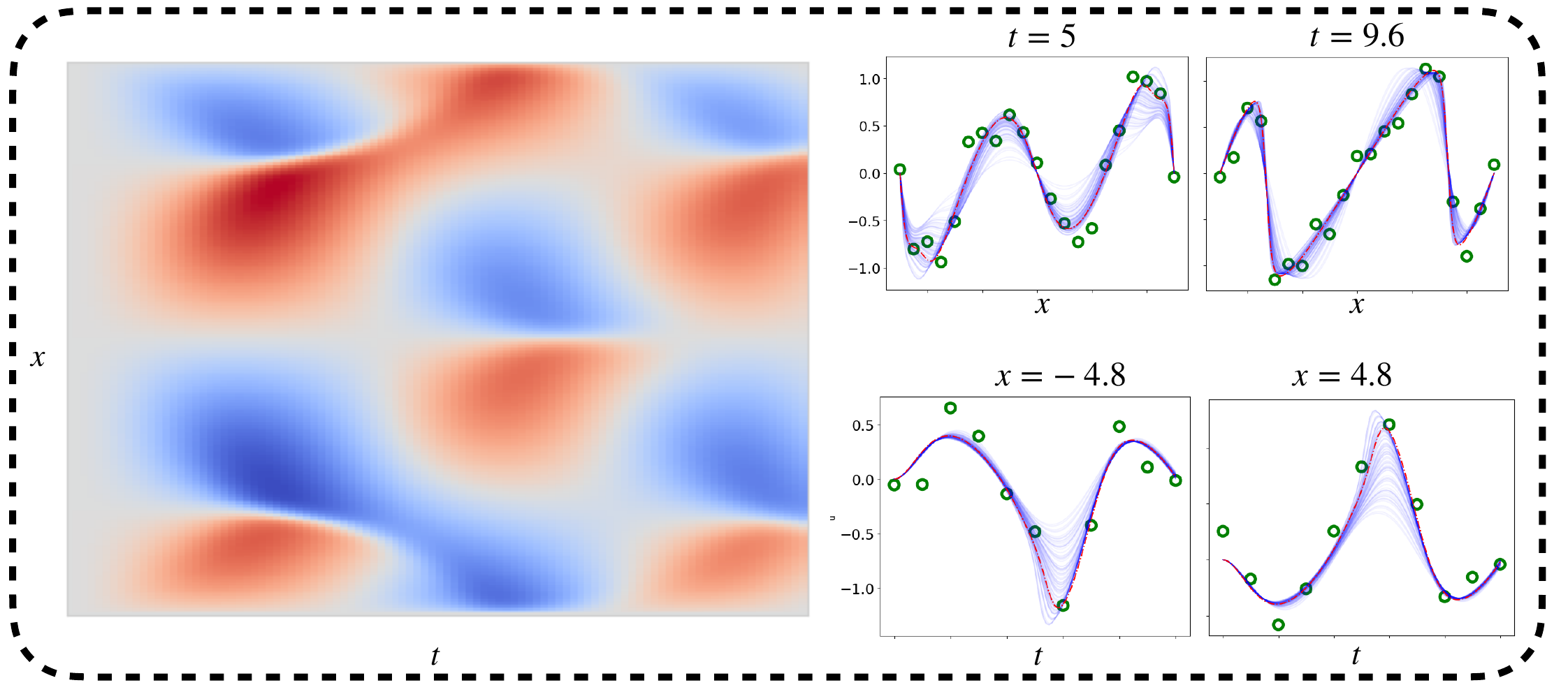}
   \caption{Additional UQ results for PDE; the black box shows the cross section UQ results for Burgers' equation with source.}
   \label{fig:addUQPDE3}
 \end{figure}

\begin{figure}[H]
  \centering
  \includegraphics[width=1.0\textwidth]{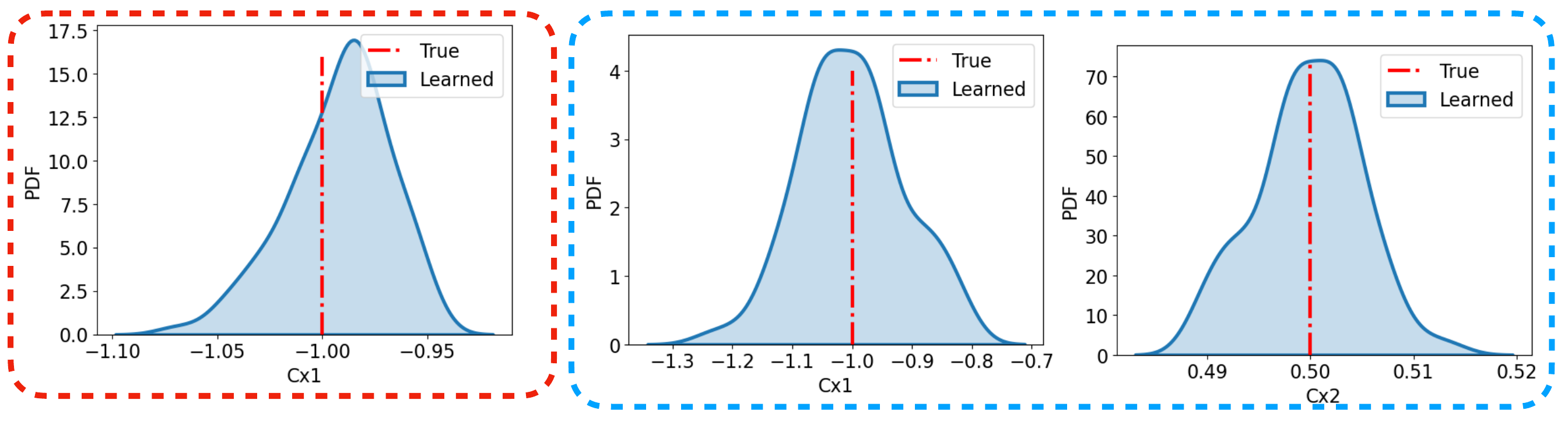}
  \caption{Additional discovery results for PDE systems; the red box shows results coefficients distribution for the advection equation, and the blue box shows the result for Burgers' equation.}
  \label{fig:addPDE}
 \end{figure}

\begin{figure}[H]
  \includegraphics[width=0.5\textwidth]{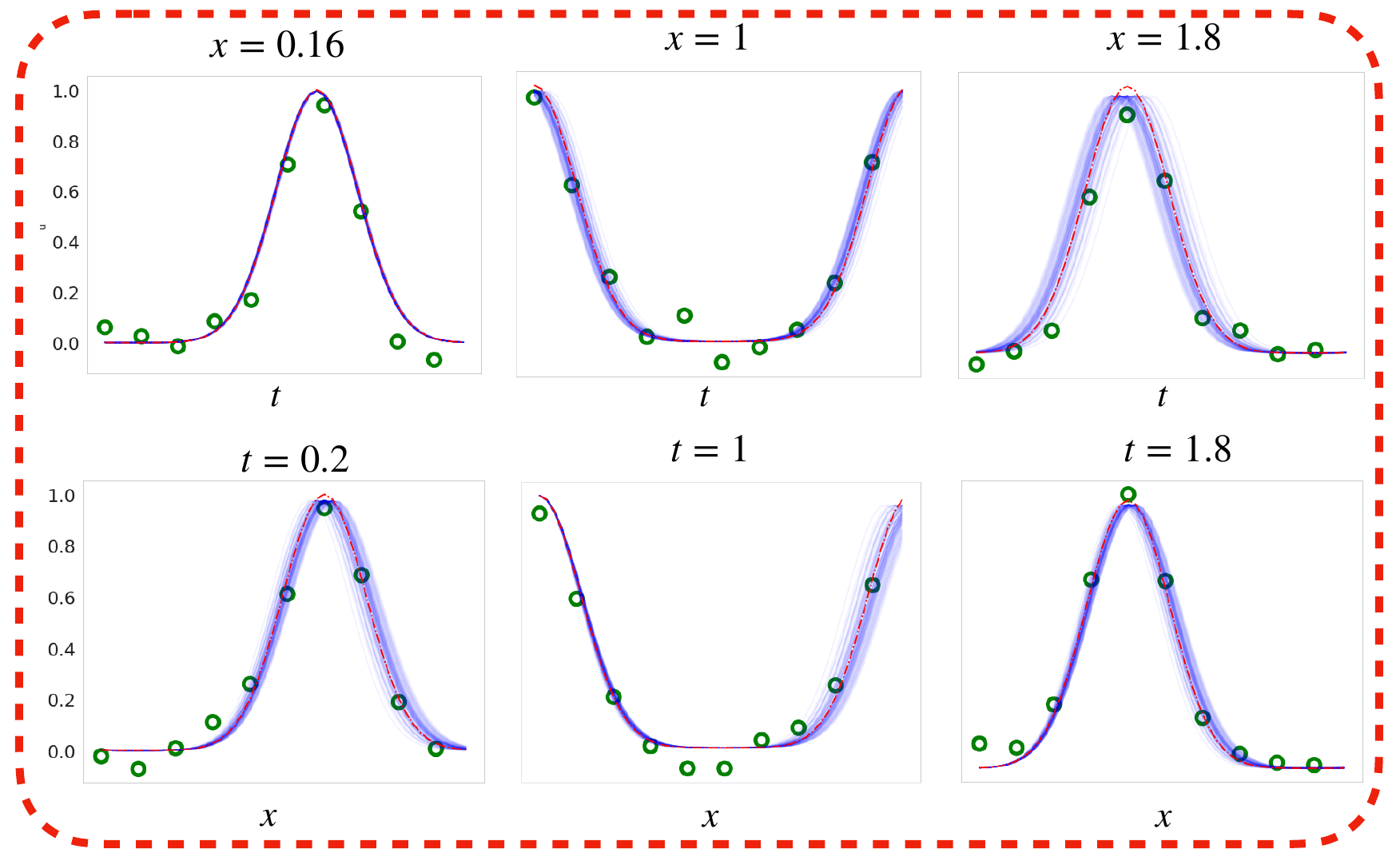}
  \hfill
  \includegraphics[width=0.45\textwidth]{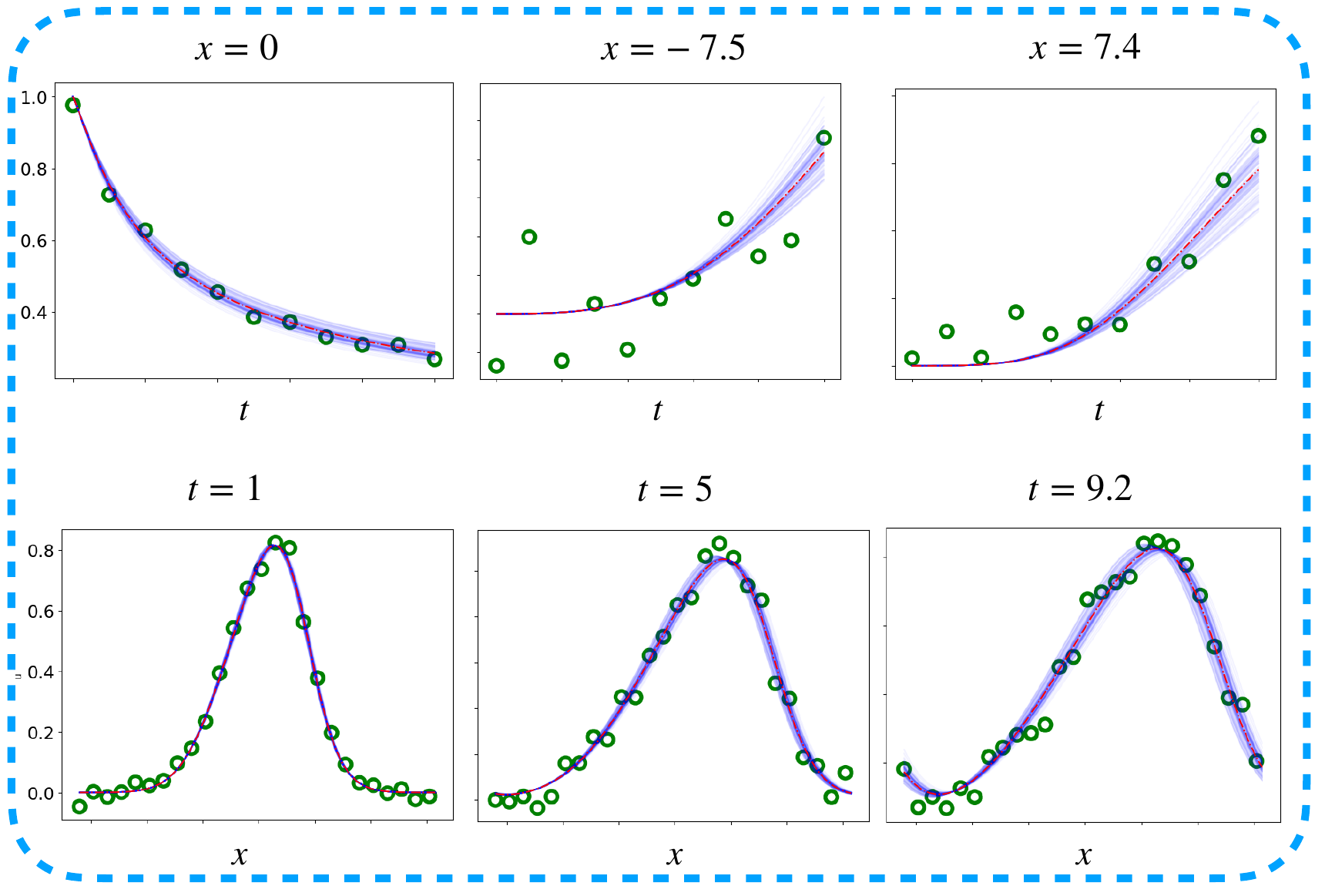}
  \caption{Additional UQ results PDE; the red box shows the cross section UQ results for Advection equation. The blue box shows the cross section UQ results for Burgers' equation.}
  \label{fig:addUQPDE}
 \end{figure}

In this section, we attached the qualitative result for PDE discovery and the uncertainty quantification.
The contour plot and the cross section result are shown in Fig~\ref{fig:PDE} (for advection equation and Burgers equation) and Fig.~\ref{fig:addUQPDE3} (for Burgers equation with source). The analytical form of the mentioned PDEs are listed in Tab.~\ref{tab:aupde}. Probability distribution for the PDE coefficient are shown in Fig.~\ref{fig:addPDE}. Additional UQ prediction result for advection and Burgers' equation are shown in and Fig.~\ref{fig:addUQPDE}.

  
 
 \newpage
 \subsection{Additional Discovery result}
 In this section, we list the full table that includes all the experiments made for current work, as attached in Tab.~\ref{tab:FullODEcp1}.
 \begin{table}[H]
  \caption{ODE and PDE discovery comparison}
  \label{tab:FullODEcp1}
  \vspace{6pt}
  \centering
  \footnotesize
  \begin{tabular}{lllllll}
    \toprule
    Name     &{$\textbf{rmse}$}($0\%$)    &{$\textbf{rmse}$}($1\%$) &{$\textbf{rmse}$} (large\tablefootnote{Large noise for different cases: Van der Pol: $5\%$, Lorenz 96: $10\%$, Advection: $20\%$, Burgers: $10\%$, Burgers' with source: $20\%$, Heat: $15\%$, Poisson: $5\%$})& {$\mathbf{M_P}$} &  {$\mathbf{M_R}$\tablefootnote{$\mathbf{M_P},\mathbf{M_R}$ are only reported for the largest noise cases}}&{Training Cost\tablefootnote{All cases are running on a Nvidia 2070 Ti GPU card}}\\
    \midrule
    \multicolumn{7}{c}{Van der Pol Oscillator}
    \\
    \cmidrule(r){1-7}
    \textbf{BSL(Ours)} & $\mathbf{0.2}$  &$2.82$&$\mathbf{18.04}$& {$\mathbf{1}$} &{$\mathbf{1}$}& {$\sim 133(+3)s$}   \\
    PINN-SR&$\text{Fail}$\tablefootnote{Fail means failure in discovery of the parsimonious ODE/PDE forms.}&$\text{Fail}$&$\text{Fail}$&{$0.214$}&{$0.75$}&{$\sim 1213s$}\\
    SINDy     & $1.0$       &$\mathbf{1.93}$&$\text{Fail}$& {$0.267$}&{$1.0$}& {$\sim 10s$} \\
    RVM     & {$1.0$}       &{$2.54$}&{$27.46$}& {$\mathbf{1}$}&{$\mathbf{1}$}& {$\sim 10s$} \\
    
    \toprule
    \multicolumn{7}{c}{Lorenz 96}                   \\
    \midrule
    \textbf{BSL(Ours)} & $\mathbf{0.269}$  &$1.47$&$\mathbf{13.0}$& {$\mathbf{1}$} &{$\mathbf{1}$}    &{$\sim1654(+438)s$}\\
    PINN-SR&$\text{Fail}$&$\text{Fail}$&$\text{Fail}$&{$0.5$}&{$0.22$}&{$\sim10788s$}\\
    SINDy     & $0.4$       &$0.64$&$\text{Fail}$ & {$0.75$} & {$\mathbf{1}$} & {$\sim 10s$}\\
    RVM     & {$0.4$}       &{$\mathbf{0.6}$}&{$49.7$}& {$\mathbf{1}$}&{$\mathbf{1}$}& {$\sim 25s$} \\
    \toprule
    \multicolumn{7}{c}{Advection Equation}                   \\
    \midrule
    \textbf{BSL(Ours)} & $\mathbf{0.26}$  &$\mathbf{1}$&$\mathbf{1.9}$& {$\mathbf{1}$} &{$\mathbf{1}$}    &{$\sim 946(+233)s$}\\
    PINN-SR & $5.9$ &$4.5$&$30.4$ &{$\mathbf{1}$} &{$\mathbf{1}$}&{$\sim 650s$}\\
    SINDy     & $2.3$       &$8.2$&$38.9$& ${\mathbf{1}}$&${\mathbf{1}}$& {$\sim 10s$} \\
    RVM     & {$0.77$}       &{$6.76$}&{$\text{Fail}$}& ${0.2}$&${\mathbf{1}}$& {$\sim 4s$} \\
    \toprule
    \multicolumn{7}{c}{Burgers' Equation}                   \\
    \midrule
    \textbf{BSL(Ours)} & $3.62$  &$4.13$&$\mathbf{6.38}$& ${\mathbf{1}}$ &${\mathbf{1}}$ & {$\sim 117(+74)s$}  \\
    PINN-SR & $10.2$ &$\mathbf{3.3}$&$10.3$ &${\mathbf{1}}$      &${\mathbf{1}}$&{$\sim 512s$}\\
    SINDy     & $0.826$       &Fail&Fail & ${\mathbf{1}}$ & ${0.5}$& {$\sim 10s$} \\
    RVM     & {$\mathbf{0.754}$}       &{$
    \text{Fail}$}&{$\text{Fail}$}& ${0.1429}$&${0.5}$& {$\sim 4s$} \\
    \midrule
  

    Name &{$\textbf{rmse}$}($0\%$)&{$\textbf{rmse}$}($0.1\%$)&{$\textbf{rmse}$} (large)& {$\mathbf{M_P}$}&  {$\mathbf{M_R}$} & {Training Cost}\\
    \midrule
     \multicolumn{7}{c}{Burgers' with Source}
    \\
    \cmidrule(r){1-7}
    \textbf{BSL(Ours)}&{$11$}&{$\mathbf{12.4}$}&{$\mathbf{13.4}$}& ${\mathbf{1}}$ &${\mathbf{1}}$ & {$\sim 396(+340)s$}    \\
    PINN-SR&{$\mathbf{10.5}$}&{$15$}&{$34.6$}&${1}$&${1}$&{$\sim 600s$}\\
    SINDy  &{$26.2$}&{$\text{Fail}$}&{$\text{Fail}$}& ${\mathbf{1}}$&${0.33}$  & {$\sim 10s$}\\
    RVM&{$27.6$}&{$\text{Fail}$}&{$\text{Fail}$}& ${0.5}$&${0.67}$  & {$\sim 10s$}\\
    \toprule
    \multicolumn{7}{c}{Heat Equation}
    \\
    \midrule
    \textbf{BSL(Ours)}&{$19$}&{$19$}&{$\mathbf{38.9}$}& ${\mathbf{1}}$ &${\mathbf{1}}$ & {$\sim 71(+8)s$}    \\
    PINN-SR&{$\text{Fail}$}&{$\text{Fail}$}&{$\text{Fail}$}&{$0$} &{$0$}&{$\sim 285s$}\\
    SINDy  &{$\mathbf{1.9}$}&{$\mathbf{17}$}&{$\text{Fail}$}& ${0.25}$ &${1}$  & {$\sim 10s$}\\
    RVM  &{$2.8$}&{$6$}&{$\text{Fail}$}& ${0}$ &${0}$  & {$\sim 6s$}\\
    \toprule
    \multicolumn{7}{c}{Poisson Equation}
    \\
    \midrule
    \textbf{BSL(Ours)}&{$\mathbf{1.18\times10^{-2}}$}&{$\mathbf{0.133}$}&{$\mathbf{16.7}$}& {${\mathbf{1}}$} &${\mathbf{1}}$ & {$\sim 92(+9)s$}    \\
    PINN-SR&{$\text{Fail}$}&{$\text{Fail}$}&{$\text{Fail}$}&${0}$ &${0}$&{$\sim 3737s$}\\
    SINDy  &{$1.15$}&{$\text{87}$}&{$\text{962}$}& ${\mathbf{1}}$&${\mathbf{1}}$  & {$\sim 10s$}\\
    RVM  &{$1.15$}&{$232$}&{$968$}& ${\mathbf{1}}$&${\textbf{1}}$  & {$\sim 10s$}\\
    \bottomrule
  \end{tabular}
 \end{table} 

 \newpage
 \subsection{Analytical forms of the discovered system}
\begin{table}[H]
  \caption{Analytical forms of ODE}
  \label{tab:aode}
  \centering
  \scriptsize
  \begin{tabular}{cc}
    \toprule
    
    Name &$\epsilon$ (large)\\
    \midrule
    \multicolumn{2}{c}{Vander Pol Oscillator}
    \\
    \cmidrule(r){1-2}
    True&$\frac{dx}{dt}=y$,$\frac{dy}{dt} = -x-0.5x^2y+0.5y$\\
    \midrule
    BSL(Ours)& $\frac{dx}{dt}=1.0096(\pm 0.024)y$,$\frac{dy}{dt} = -0.9858(\pm 0.037) x-0.4801(\pm 0.114)x^2y+0.4889(\pm 0.111)y$  \\
    \midrule
    PINN-SR&$\frac{dx}{dt} = 1.3079-0.1151x+0.9982y+0.1939x^2-0.4101xy+0.8559y^2-0.3577x^3$\\
    &$+0.6764x^2y-0.1207xy^2+0.6261y^3$,\\
    &$\frac{dy}{dt}=-0.6718x+1.6035y+1.6339y^2+0.4803y^3$\\
    \midrule
    SINDy &$\frac{dx}{dt}=0.1197+0.169x+0.9975y-0.0292x^2+0.0232xy-0.0274y^2$\\
    &$-0.0393x^3+0.0208x^2y-0.0463xy^2$,\\
    &$\frac{dy}{dt} = -1.0909x+0.149y+0.0294x^3-0.4x^2y-0.0356xy^2+0.0762y^3$\\
    \midrule
    RVM&$\frac{dx}{dt}=0.9957(\pm0.073)y$,$\frac{dy}{dt} = -0.9943(\pm0.0597)x-0.4777(\pm0.0809)x^2y+0.4714(\pm0.1204)y$\\
    \midrule
    \multicolumn{2}{c}{Lorenz 96}                   \\
    \midrule
    True&$\frac{dX_1}{dt} = (X_2-X_5)X_6-X_1+8$, $\frac{dX_2}{dt} = (X_3-X_6)X_1-X_2+8$,\\
    &$\frac{dX_3}{dt} = (X_4-X_1)X_2-X_3+8$, $\frac{dX_4}{dt} = (X_5-X_2)X_3-X_4+8$, \\
    &$\frac{dX_5}{dt} = (X_6-X_3)X_4-X_5+8$, $\frac{dX_6}{dt} = (X_1-X_4)X_5-X_6+8$\\
    \midrule
    BSL(Ours)& $\frac{dX_1}{dt} = 1.0033(\pm2.37\times10^{-4})X_2X_6-0.9926(\pm2.03\times10^{-4})X_5X_6$,\\
    &$-0.9991(\pm9.04\times10^{-4})X_1+7.9773(\pm2.08\times10^{-2})$,\\ &$\frac{dX_2}{dt} = 0.9963(\pm5.54\times10^{-5})X_1X_3-0.9942(\pm2.47\times10^{-4})X_1X_6$\\
    &$-1.0054(\pm2.31\times10^{-4})X_2+7.8719(\pm7.05\times10^{-4})$,\\
    &$\frac{dX_3}{dt} = 1.0106(\pm{2.19\times10^{-4}})X_2X_4-1.0029(\pm{2.03\times10^{-4}})X_1X_2$\\
    &$-0.9979(\pm5.3\times10^{-4})X_3+7.9938(\pm{1.48\times10^{-2}})$,\\ 
    &$\frac{dX_4}{dt} = 1.0067(\pm1.01\times10^{-4})X_3X_5-1.0103(\pm9.09\times10^{-5})X_2X_3$\\
    &$-0.9926(\pm2.36\times10^{-4})X_4+8.055(\pm{2.41\times10^{-3}})$, \\
    &$\frac{dX_5}{dt} = 0.9953\pm(4.39\times10^{-4})X_4X_6-0.9922\pm(3.88\times10^{-4})X_3X_4$\\
    &$-1.0072\pm(4.1\times10^{-4})X_5+7.8538(\pm{2.1\times10^{-3}})$,\\ 
    &$\frac{dX_6}{dt} = 1.0095(\pm2.71\times10^{-4})X_1X_5-0.9967(\pm2.83\times10^{-4})X_4X_5$\\
    &$-0.9772(\pm2.46\times10^{-4})X_6+7.9964(\pm4.15\times10^{-2})$\\
    \midrule
    PINN-SR& N/A\\
    \midrule
    SINDy &$\frac{dX_1}{dt} = 8.0985-0.9423X_1-0.1007X_4-0.1659X_6+0.9582X_2X_6-0.944X_5X_6$\\ 
    &$\frac{dX_2}{dt} = 7.9372-0.1268X_1-0.9607X_2+0.9636X_1X_3-0.9558X_1X_6$\\
    &$\frac{dX_3}{dt} = 8.1523-0.0983X_1-0.1652X_2-0.9952X_3-0.9541X_1X_2+0.9651X_2X_4$\\ 
    &$\frac{dX_4}{dt} = 7.4958-0.9754X_4+0.1515X_5-0.973X_2X_3+0.9206X_3X_5$, \\
    &$\frac{dX_5}{dt} = 7.8556-0.1146X_4-0.9457X_5-0.9559X_3X_4+1.0023X_4X_6$\\
    &$\frac{dX_6}{dt} = 7.6026+0.0934X_2-0.927X_6+0.9711X_1X_5-0.9754X_4X_5$\\
    \midrule
    RVM& $\frac{dX_1}{dt} = 0.9613(\pm1.21)X_2X_6-0.962(\pm1.27)X_5X_6$,\\
    &$-0.8983(\pm2.67)X_1+7.518(\pm5.88)$,\\ &$\frac{dX_2}{dt} = 0.9578(\pm1.31)X_1X_3-0.9693(\pm1.24)X_1X_6$\\
    &$-0.9217(\pm2.8)X_2+7.6442(\pm6.1575)$,\\
    &$\frac{dX_3}{dt} = 0.9535(\pm{1.48})X_2X_4-0.9803(\pm{1.38})X_1X_2$\\
    &$-0.9468(\pm2.96)X_3+7.5323(\pm6.56)$,\\ 
    &$\frac{dX_4}{dt} = 0.9365(\pm1.24)X_3X_5-0.9748(\pm1.15)X_2X_3$\\
    &$-0.9114(\pm2.59)X_4+7.67(\pm5.49)$, \\
    &$\frac{dX_5}{dt} = 0.9981\pm(1.26)X_4X_6-0.9667\pm(1.2)X_3X_4$\\
    &$-0.9216\pm(2.75)X_5+7.6146(\pm5.96)$,\\ 
    &$\frac{dX_6}{dt} = 0.9677(\pm1.42)X_1X_5-0.9757(\pm1.38)X_4X_5$\\
    &$-0.8986(\pm2.86)X_6+7.6657(\pm6.57)$\\
    \bottomrule
  \end{tabular}
 \end{table}
\newpage
 \begin{table}[H]
  \caption{Analytical forms of unsteady PDE}
  \label{tab:aupde}
  \centering
  \footnotesize
  \begin{tabular}{cc}
    \toprule

    Name &$\epsilon$ (large)\\
    \midrule
    \multicolumn{2}{c}{Advection Equation}
    \\
    \cmidrule(r){1-2}
    True&$u_t=-u_x$\\
    BSL(Ours)&   $u_t = -0.9988(\pm0.024)u_x$\\
    PINN-SR&$u_t=-0.997u_x$\\
    SINDy &$u_t=-0.9961u_x$\\
    RVM&$u_t = -0.5148(\pm0.106)u_x-0.9797(\pm0.384)uu_x$\\
    &$-0.018(\pm0.208)u^2u_x-0.075(\pm0.148)u^2u_{xxx}+0.049(\pm0.111)u^3u_{xxx}$\\
    \midrule
    \multicolumn{2}{c}{Burgers' Equation}                   \\
    \midrule
    True&$u_t = -uu_x+0.5u_{xx}$\\
    BSL(Ours)&  $u_t = -0.9929(\pm0.086)uu_x+0.4993(\pm0.005)u_{xx}$ \\
    PINN-SR&$u_{t} = -1.0103uu_{x}+0.5051u_{xx}$\\
    SINDy &$u_t=-0.8179uu_x$\\
    RVM&$-0.0809(\pm0.041)u_x-1.6684(\pm0.2618)uu_x+4.1835(\pm0.571)u^2u_x$\\
    &$-3.9068(\pm0.391)u^3u_x+0.1916(\pm0.064)uu_{xx}$\\
    &$-1.0314(\pm0.197)u^2u_{xx}+1.5504(0.156)u^3u_{xx}$\\
    \midrule
    \multicolumn{2}{c}{Burgers' Equation with Source}                   \\
    \midrule
    True&$u_t = -uu_x+0.1u_{xx}+sin(x)sin(t)$\\
    BSL(Ours)&  $-0.9882(\pm0.246)uu_x+0.105(\pm0.022)u_{xx}+0.9859(\pm0.005)sin(x)sin(t)$ \\
    PINN-SR&$u_t = -0.9576uu_x+0.1168u_{xx}+1.0179sin(x)sin(t)$\\
    SINDy &$u_t =0.8052sin(x)sin(t)$\\
    RVM&$-0.0234(\pm0.145)uu_x+0.8318(\pm0.142)sin(x)sin(t)$\\
    &$-0.0789(\pm0.105)sin(x)+0.3558(\pm0.156)sin(x)cos(t)$\\
    \bottomrule
  \end{tabular}
 \end{table}

 \begin{table}[H]
  \caption{Analytical forms of steady PDE}
  \label{tab:aspde}
  \centering
  \footnotesize
  \begin{tabular}{cc}
    \toprule
    Name &$\epsilon$ (large)\\
    \midrule
    \multicolumn{2}{c}{Heat Equation}
    \\
    \cmidrule(r){1-2}
    True&$u_{yy}=-u_{xx}$\\
    BSL(Ours)&  $u_{yy}=-0.9611(\pm0.059)u_{xx}$ \\
    PINN-SR&$u_{yy}=0.5544u_x$\\
    SINDy &$u_{yy}=-0.069u_{xx}+13.8988uu_x-19.493u_x+0.2468uu_{xx}$\\
    RVM&$u_{yy}=-7.5861(\pm202.6)u_{x}$\\
    \midrule
    \multicolumn{2}{c}{Poisson Equation}                   \\
    \midrule
    True&$u_{yy} = -u_{xx}-sin(x)sin(y)$\\
    BSL(Ours)&  $u_{yy} = -0.9788(\pm8.75\times10^{-4})u_{xx}-0.9897(\pm4.63\times10^{-4})sin(x)sin(y)$\\
    PINN-SR&$u_{yy} = 0.13611752uu_x+0.29748484uu_{xxx}$\\
    SINDy &$u_{yy}=0.2316u_{xx}-0.4221sin(x)sin(y)$\\
    RVM&$u_{yy}=0.2284(\pm0.007)u_{xx}-0.3954(\pm0.01)sin(x)sin(y)$\\
    \bottomrule
  \end{tabular}
 \end{table}
 
  \subsection{Implementation detail for the spline}
  {
  In the current work, we apply direct tensor product to extend spline for solving spatial-temporal field and the relevant statistics are attached in Tab.~\ref{tab:tp_spline}, where the numerber of control points (trainable weights) $\theta$ in 1-d is marked by red and the total number of control points for the 2-d scenario is listed in the last column. We only store the non-zero elements for two-dimensional basis to leverage the sparsity (local support) of the spline. However, we must claim that it is \emph{not} an optimal way to extend spline for higher spatial dimensions. In that case, a spline kernel can be defined and the tensor-product is only processed in the subdomain, as shown in~\cite{fey2018splinecnn}.}
  \begin{table}[H]
  \caption{Direct tensor-product spline for PDE}
  \label{tab:tp_spline}
  \centering
  \footnotesize
  \begin{tabular}{ccccc}
    \toprule
    basis $x$ &basis $t(y)$ & basis$(x,t(y))$&sparsity&trainable params\\
    \midrule
    \multicolumn{5}{c}{Advection Equation}
    \\
    \cmidrule(r){1-5}
    $50\times{\color{red}{54}}$&$50\times{\color{red}54}$&$62500$&$0.0086$&$2916$\\
    \midrule
    \multicolumn{5}{c}{Burgers' Equation}
    \\
    \cmidrule(r){1-5}
    $128\times{\color{red}13}$&$101\times{\color{red}19}$&$92872$&$0.058$&$247$\\
    \midrule
    \multicolumn{5}{c}{Burgers' Equation with Source}
    \\
    \cmidrule(r){1-5}
    $201\times{\color{red}103}$&$101\times{\color{red}103}$&$251415$&$0.0012$&$10609$\\
    \midrule
    \multicolumn{5}{c}{Heat Equation}
    \\
    \cmidrule(r){1-5}
    $51{\color{red}\times11}$&$51\times{\color{red}11}$&$41209$&$0.1309$&$121$\\
    \midrule
    \multicolumn{5}{c}{Poisson}
    \\
    \cmidrule(r){1-5}
    $101\times{\color{red}53}$&$101\times{\color{red}53}$&$162409$&$0.0057$&$2809$\\

    \bottomrule
  \end{tabular}
 \end{table}
 \subsection{Relevant Terminologies}
  \begin{table}[H]
  \caption{Terminologies}
  \label{tab:terms}
  \centering
  \scriptsize
  \begin{tabular}{ccc}
    \toprule
    Name & Symbol &Explanation\\
    \midrule
    Control points& $\mathbf{\theta}$&Trainable weight $\theta$ for spline basis.\\
    \midrule
    Knots &$\mathbf{\tau}_s$&Location of control points.\\
    \midrule
    Measurement points& &Sparse spatio-temporal points with labels.\\
    \midrule
    Collocation points& &Dense spatio-temporal points without labels.\\
    \midrule
    &$\mathbf{N_m}$&Spline basis evaluated at measurement points.\\
    \midrule
    &$\mathbf{N_c}$&Spline basis evaluated at collocation points.\\
    \midrule
    Library candidates& $\mathbf{\Phi}$&A collection of polynomial terms that \\
    & &the system identification algorithm can choose parsimonious terms from it. \\
    & & e.g., $\{x,y,x^2y,...\}$ (for ODE) or $\{u,uu_x,u_{xx}...\}$ for (PDE).\\
    \midrule
    ADO iteration& &Alternating direction optimization to update the trainable parameters \\
    & &including control points $\bold{\theta}$, weight of library candidates $\mathbf{W}$ and covariance matrices.\\
    \midrule
    Aleatoric Uncertainty&&Due to intrinsic randomness by nature, which is irreducible.  \\
    \midrule
    Epistemic Uncertainty&& Because of a lack of knowledge, which can be reduced by adding more information.\\
    \bottomrule
    \end{tabular}
\end{table}
 \subsection{Real world application: predator-prey system}
\begin{figure}[H]
  \includegraphics[width=0.95\textwidth]{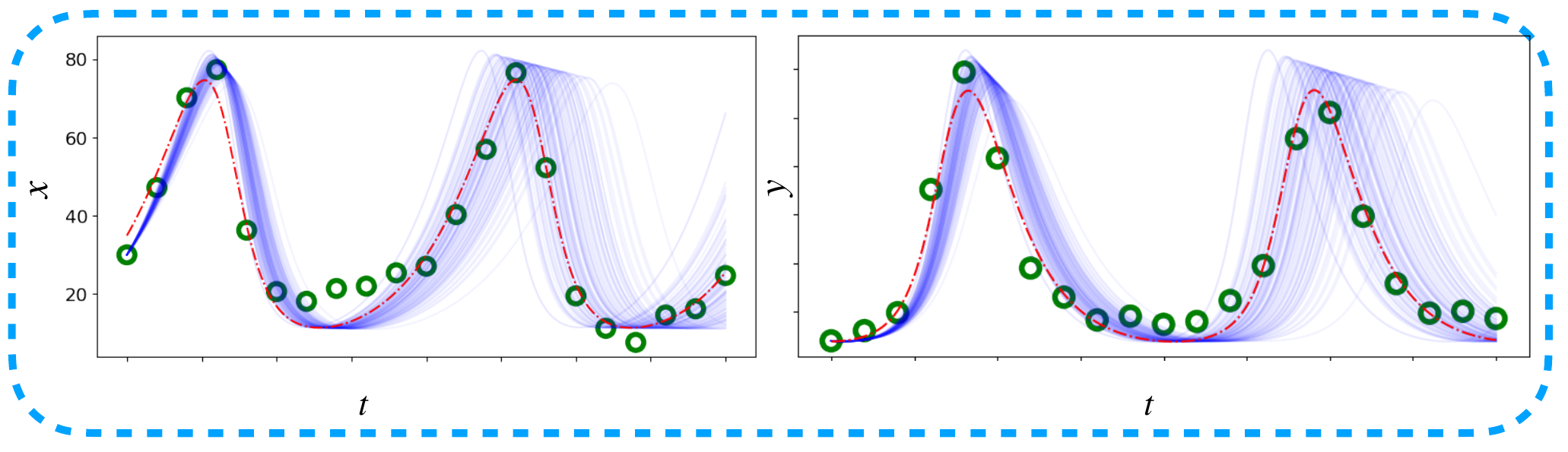}

  \caption{UQ results for predator-prey system; left sub-fig shows the result for state $x$,the hares population; the right sub-fig shows the result for state $y$, the lynx population. The green dot is the sparse and biased measurement data. The red line is the reference model and the blue curves are the ensemble predictions from our proposed model.}
  \label{fig:UQpp}
 \end{figure}

 In this section, we would test our proposed BSL model on one real-world case, predator-prey system. The real data set is obtained online and it depicts the population of hares and lynx from 1900 to 1920 from Hudson Bay Company. The data is presented in Tab.~\ref{tab:realdata}: The reference governing equation by mathematical analysis is:
 \begin{align}
 \label{eq:realref}
 \frac{dx}{dt} &= 0.4807x-0.0248xy\\
 \frac{dy}{dt} &=
 -0.9272y+0.0276xy
 \end{align}
 We test the 4 methods on this data set and the result can be found in Tab~\ref{tab:discrealODE}
 .We have made assumptions about constructing the libraries. We assume the predator (lynx) only feeds on the prey (hares). Meantime, the prey (hares) only has one predator (lynx). Therefore, the change rate of these two species can only depend on themselves ($x$,$y$) and some higher order correlations between them ($xy,x^2y,xy^2$). The discovered forms of the 4 methods are listed in Tab.~\ref{tab:anarealODE}. Finally, the UQ results are shown in Fig~\ref{fig:UQpp}. In short, only the proposed method work on the real sparse and noisy dataset. And the UQ prediction covers more measurement points than the reference model Eq.~\ref{eq:realref}, which helps to better explain the data set.
  \begin{table}[H]
  \caption{ODE discovery comparison}
  \label{tab:discrealODE}
  \vspace{6pt}
  \centering
  \footnotesize
  \begin{tabular}{lllll}
    \toprule
    Name     &{$\textbf{rmse}$} (large)& {$\mathbf{M_P}$} &  {$\mathbf{M_R}$}&{Training Cost}\\
    \midrule
    \multicolumn{5}{c}{Predator-prey}
    \\
    \cmidrule(r){1-5}
    \textbf{BSL(Ours)} &${\mathbf{30.4}}$& {$\mathbf{1}$} &{$\mathbf{1}$}& {$\sim 2278(+6) s$}   \\
    PINN-SR&${\text{Fail}}$&{$0.5$}&{$0.25$}&{$\sim 2988s$}\\
    SINDy   &{$\text{Fail}$}& {$0.6$}&{$0.75$}& {$\sim 10s$} \\
    RVM     &{$\text{Fail}$}& {$0.8$}&{$1$}& {$\sim 10s$} \\
    \bottomrule
    \end{tabular}
\end{table}
 
 \begin{table}[H]
  \caption{Analytical forms of ODE}
  \label{tab:anarealODE}
  \vspace{6pt}
  \centering
  \scriptsize
  \begin{tabular}{cc}
    \toprule
    Name &$\epsilon$ (large)\\
    \midrule
    \multicolumn{2}{c}{Predator-prey (Lotka-Volterra)}
    \\
    \cmidrule(r){1-2}
    True&$\frac{dx}{dt}=0.4807x-0.0248xy$,$\frac{dy}{dt} = -0.9272y+0.0276xy$\\
    \midrule
    BSL(Ours)& $\frac{dx}{dt}=-0.5124(\pm0.028)x-0.0266(\pm8.72\times10^{-4})xy$\\
    &$\frac{dy}{dt} =-0.9258(\pm0.065)y+0.0279(\pm1.47\times10^{-3})xy $  \\
  
    \midrule
    PINN-SR&$\frac{dx}{dt} = -13.9238y$\\
    &$\frac{dy}{dt}=-0.1144y$\\
    \midrule
    SINDy &$\frac{dx}{dt}=0.5813x-0.0261xy$,\\
    &$\frac{dy}{dt} = 0.2549x-0.2702y$\\
    \midrule
    RVM&$\frac{dx}{dt}=0.5732(\pm0.6488)x-0.2386(\pm0.4643)y-0.0253(\pm0.1432)xy$,\\
    &$\frac{dy}{dt} =-0.8018(\pm0.9459)y+0.0226(\pm0.1481)xy $\\
    \bottomrule
    \end{tabular}
\end{table}

\begin{table}[H]
  \caption{Lynx-Hares population}
  \label{tab:realdata}
  \vspace{6pt}
  \centering
  \scriptsize
  \begin{tabular}{ccc}
    \toprule
    {Year} &{Hares($\times 1000$)} & {Lynx($\times 1000$)}\\
    \midrule
    {1900}&{30}&{4}\\
    \midrule
    {1901}&{47.2}&{6.1}\\
    \midrule
    {1902}&{70.2}&{9.8}\\
    \midrule
    {1903}&{77.4}&{35.2}\\
    \midrule
    {1904}&{36.3}&{59.4}\\
    \midrule
    {1905}&{20.6}&{41.7}\\
    \midrule
    {1906}&{18.1}&{19}\\
    \midrule
    {1907}&{21.4}&{13}\\
    \midrule
    {1908}&{22}&{8.3}\\
    \midrule
    {1909}&{25.4}&{9.1}\\
    \midrule
    {1910}&{27.1}&{7.4}\\
    \midrule
    {1911}&{40.3}&{8}\\
    \midrule
    {1912}&{57}&{12.3}\\
    \midrule
    {1913}&{76.6}&{19.5}\\
    \midrule
    {1914}&{52.3}&{45.7}\\
    \midrule
    {1915}&{19.5}&{51.1}\\
    \midrule
    {1916}&{11.2}&{29.7}\\
    \midrule
    {1917}&{7.6}&{15.8}\\
    \midrule
    {1918}&{14.6}&{9.7}\\
    \midrule
    {1919}&{16.2}&{10.1}\\
    \midrule
    {1920}&{24.7}&{8.6}\\
    \bottomrule
 \end{tabular}
\end{table}
 \subsection{Experiment on different smoothing algorithms}
 This section we will test the effect of different smoothing algorithms and its impact to the SINDy result. These methods include smoothing based polynomial interpolation, convolutional smoother, smoothing with Tikhonov regularization and smoothing with spline fitting. We performed comparison of SINDy with these smoothing methods on a representative ODE system (Van der Pol system) and PDE system (Poisson equation) studied in this work.  As shown in the Table~\ref{tab:smoothmetric} and Table~\ref{tab:smooth_eq} below, when the data noise is above $5\%$, although data is preprocessed using smoothing and uniform resampling, none of these method work. Basically, the SINDy still failed to discover the correct model forms with different smoothing schemes, and the identified systems are different from the true. In contrast, our proposed approach is very robust and superior to handling corrupted data, thanks to the spline learning in Bayesian settings.
   \begin{table}[h]
  \caption{ODE and PDE discovery comparison}
  \label{tab:smoothmetric}
  \vspace{6pt}
  \centering
  \footnotesize
  \begin{tabular}{lllll}
    \toprule
    Name     &{$\textbf{rmse}(\epsilon = 5\%)$} & {$\mathbf{M_P}$} &  {$\mathbf{M_R}$}&{Training Cost}\\
    \midrule
    \multicolumn{5}{c}{Van der Pol Oscillator}
    \\
    \cmidrule(r){1-5}
    \textbf{BSL(Ours)} &{$\mathbf{18.04}$}& {$\mathbf{1}$} &{$\mathbf{1}$}& {$\sim 133(+3)s$}  \\
    SINDy(No smoother)     &{$\text{Fail}$}& {$0.33$}&{$\mathbf{1}$}& {$\sim 10s$} \\
    SINDy(Poly)   &{$\text{Fail}$}& {$0.6$}&{$0.75$}& {$\sim 10s$} \\
    
    SINDy(Conv)     &{$\text{Fail}$}& {$0.4$}&{$\mathbf{1}$}& {$\sim 10s$} \\
    SINDy(Tikhonov)     &{$\text{Fail}$}& {$0.21$}&{$\mathbf{1}$}& {$\sim 10s$} \\
    SINDy(Spline)
    &{$\text{Fail}$}&{$0.25$}&{$1$}&{$\sim 10s$}\\
    \midrule
    \multicolumn{5}{c}{Poisson Equation}\\
    \midrule
    \textbf{BSL(Ours)}&{$\mathbf{16.7}$}& {${\mathbf{1}}$} &${\mathbf{1}}$ & {$\sim 92(+9)s$}    \\
    SINDy(No smoother)   &{$\text{Fail}$}& {$0.5$}&{$\mathbf{1}$}& {$\sim 10s$} \\
    SINDy(Poly)  &{$962$}& ${\mathbf{1}}$&${\mathbf{1}}$  & {$\sim 10s$}\\
    
    SINDy(Conv)   &{$\text{Fail}$}& {$0.66$}&{$\mathbf{1}$}& {$\sim 10s$} \\
    SINDy(Tikhonov)   &{$\text{Fail}$}& {$0.5$}&{$\mathbf{1}$}& {$\sim 10s$} \\
    SINDy(Spline)
    &{$\text{Fail}$}& {$0.5$}&{$\mathbf{1}$}& {$\sim 10s$} \\
    \bottomrule
    \end{tabular}
\end{table}
  \begin{table}[h]
  {
  \caption{Smoothing algorithm effects}
  \label{tab:smooth_eq}
  \centering
  \footnotesize
  \begin{tabular}{cc}
    \toprule

    Name & Analytical form\\
    \midrule
    \multicolumn{2}{c}{Van der Pol Oscillator}\\
    \midrule
    True & $\frac{dx}{dt} = y$\\
    & $\frac{dy}{dt} = -x-0.5x^2y+0.5y$
    \\
    \midrule
    SINDy(No smoother) &$\frac{dx}{dt}=0.0277-0.217x+1.4y-0.025x^2$\\
    &$+0.0542x^3-0.1285x^2y+0.1039xy^2-0.0896y^3$,\\
    &$\frac{dy}{dt} = -0.9835x+0.3462y-0.4675x^2y+0.0325y^3$\\
    \midrule
    SINDy(Poly) &$\frac{dx}{dt}=0.1197+0.169x+0.9975y-0.0292x^2$\\
    &$+0.0232xy-0.0274y^2-0.0393x^3+0.0208x^2y-0.0463xy^2$,\\
    &$\frac{dy}{dt} = -1.0909x+0.149y+0.0294x^3-0.4x^2y-0.0356xy^2+0.0762y^3$\\
    \midrule
    SINDy(Conv) &$\frac{dx}{dt}=0.1518+0.9978y-0.0486x^2+0.0274xy-0.0327y^2$,\\
    &$\frac{dy}{dt} = -1.1154x-0.2143y+0.0348x^3-0.4374x^2y+0.0613y^3$\\
    \midrule
    
    SINDy(Tikhonov)&$\frac{dx}{dt}$
    $=-0.2572+0.1486x+1.1299y+0.0982x^2-0.0721xy$,\\
    &$+0.0549y^2-0.0478x^2y-0.0816xy^2-0.0542y^3$\\
    &$\frac{dy}{dt} = 0.2338-1.3282x+0.1603y-0.0718x^2$\\
    &$+0.058xy-0.0564y^2+0.1069x^3-0.0842x^2y+0.1373xy^2-0.0415y^3$\\
    \midrule
    SINDy(Spline)&$\frac{dx}{dt}= 0.1916+1.5304y-0.0614x^2+0.0215xy$\\
    &$-0.0325y^2-0.155x^2y+0.0796xy^2-0.1168y^3$\\
    &$\frac{dy}{dt} = -0.35-1.0127x+0.6756y+0.0695x^2$\\
    &$+0.0741y^2-0.5493x^2y+0.0523xy^2-0.0412y^3$\\
    \midrule
    \multicolumn{2}{c}{Poisson equation}\\
    \midrule
    True&$u_{yy} = -u_{xx}-sin(x)sin(y)$\\
    \midrule
    SINDy(No smoother)
    &$u_{yy}=0.5635u_{xx}+1.683uu_{x}+0.088uu_{xx}-0.17sin(x)sin(y)$\\
    \midrule
    SINDy(Poly) &$u_{yy}=0.2316u_{xx}-0.4221sin(x)sin(y)$\\
    \midrule
    SINDy(Conv)
    &$u_{yy}=0.47u_{xx}-0.2649sin(x)sin(y)+0.073sin(x)cos(y)$\\
    \midrule
    SINDy(Tikhonov)&$u_{yy}=-0.17u_{xx}-0.087uu_{x}-0.458sin(x)sin(y)+0.02sin(x)cos(y)$\\
    \midrule
    SINDy(Spline)&$u_{yy}=0.1562u_{xx}+0.1751uu_{x}-0.3951sin(x)sin(y)-0.0911sin(x)cos(y)$\\
    
    \bottomrule
    \end{tabular}}
\end{table}


\end{document}